\documentclass[review]{elsarticle}

\usepackage{lineno,hyperref}
\modulolinenumbers[5]

\journal{Journal of \LaTeX\ Templates}

\usepackage{moreverb,url}

\usepackage{hyperref}
\usepackage{graphicx}
\usepackage[table]{xcolor}
\usepackage{multirow}
\usepackage{enumerate}
\usepackage{array,booktabs,arydshln}
\usepackage{graphicx}
\usepackage[acronym,nomain]{glossaries}
\usepackage{moreverb,url}
\usepackage{amsmath}
\usepackage[table]{xcolor}
\usepackage{multirow}
\usepackage{enumerate}
\usepackage{array,booktabs,arydshln}
\usepackage{bbm}
\usepackage{algorithm}
\usepackage{algorithmic}
\usepackage{subfig}
\usepackage{amsfonts}
\usepackage{diagbox}
\usepackage{makecell}

\DeclareMathOperator*{\argmax}{argmax}
\DeclareMathOperator*{\argmin}{argmin}
\DeclareMathOperator*{\mode}{mode}  
\newcommand{\rom}[1]
    {\MakeUppercase{\romannumeral #1}}

\begin{document}

\begin{frontmatter}

\title{APPL: Adaptive Planner Parameter Learning}

\author{Xuesu Xiao, Zizhao Wang, Zifan Xu, Bo Liu, Garrett Warnell, Gauraang Dhamankar, Anirudh Nair, and Peter Stone}
\address{Department of Computer Science, The University of Texas at Austin, Austin, Texas 78712.}
% \fntext[myfootnote]{Since 1880.}

% \author[mymainaddress,mysecondaryaddress]{Elsevier Inc}

% \author[mysecondaryaddress]{Global Customer Service\corref{mycorrespondingauthor}}
% \cortext[mycorrespondingauthor]{Corresponding author}

% \address[mymainaddress]{1600 John F Kennedy Boulevard, Philadelphia}
% \address[mysecondaryaddress]{360 Park Avenue South, New York}

\begin{abstract}
While current autonomous navigation systems allow robots to successfully drive themselves from one point to another in specific environments, they typically require extensive manual parameter re-tuning by human robotics experts in order to function in new environments.
Furthermore, even for just one complex environment, a single set of fine-tuned parameters may not work well in different regions of that environment.
These problems prohibit reliable mobile robot deployment by non-expert users.
%and prevent continual improvement as robots gather more deployment experience.
As a remedy, we propose \emph{Adaptive Planner Parameter Learning} (\textsc{appl}), a machine learning framework that can leverage non-expert human interaction via several modalities---including teleoperated demonstrations, corrective interventions, and evaluative feedback---and also unsupervised reinforcement learning to learn a \emph{parameter policy} that can dynamically adjust the parameters of classical navigation systems in response to changes in the environment.
% (e.g., inflation radius, sampling rate, optimization coefficients). 
\textsc{appl} inherits safety and explainability from classical navigation systems while also enjoying the benefits of machine learning, i.e., the ability to adapt and improve from experience.
We present a suite of individual \textsc{appl} methods and also a unifying cycle-of-learning scheme that combines all the proposed methods in a framework that can improve navigation performance through continual, iterative human interaction and simulation training.
\end{abstract}

\begin{keyword}
Mobile Robot Navigation, Machine Learning, Motion Planning
\end{keyword}

\end{frontmatter}

% \linenumbers

%%%%%%%%%%%%%%%%%%%%%%%%%%%%%%%%%%%%%%%%%%%%%%%%%%%%%%%%%%%%%%%%%%%%%%%%%%%%%%%%
\section{INTRODUCTION}
\label{sec::introduction}

While engineered systems developed over the past several decades have enabled autonomous mobile robot navigation in certain scenarios, it is typically the case that the parameters of these systems need to be adjusted for each new deployment environment.
This process is referred to as ``parameter-tuning,'' i.e., the process through which a human with expert knowledge of the underlying system and the role of each parameter---along with trial-and-error, experience, and intuition---manually finds the best set of system parameters for a new environment. 
This traditional setup (1) precludes non-expert users from optimizing the default robot system; (2) assumes a \emph{single} set of parameters will work well for all regions of a specific environment; and (3) completely abandons previously fine-tuned parameters when using a new set of parameters.

In contrast, many humans---even those with little to no robotics experience---can easily teleoperate mobile robots in new environments \cite{farkhatdinov2009user}, and even more humans are able to provide simple evaluative feedback (e.g., a numerical assessment of performance) on the robot's behavior.
These {\em in situ} end-user interactions can provide a rich source of knowledge regarding how the system should behave in deployment environments.
We hypothesize that systems that can capture and leverage these interactions will be able to quickly adapt to their deployment environments, i.e., exhibit robust and reliable navigation system performance.

In this article, we investigate this hypothesis by developing and studying a novel family of algorithms called Adaptive Planner Parameter Learning (\textsc{appl}).
In particular, we present an algorithm that can learn from human demonstrations (\textsc{appld}), an algorithm that can learn from human interventions (\textsc{appli}), and an algorithm that can learn from human-generated evaluative feedback (\textsc{apple}).
Additionally, we present an algorithm in this family that can, if available, leverage simulation experience in an unsupervised fashion using reinforcement learning (\textsc{applr}).
Importantly, the framework under which we develop \textsc{appl} does not replace existing systems for autonomous navigation, but rather {\em augments} them by providing a new learned module that adjusts only certain hyperparameters of these systems (e.g., obstacle inflation radius, sampling rate, cost function coefficients, etc.).
By adopting this approach, \textsc{appl} systems inherit the advantages of existing navigation systems (e.g., safety and explainability), while also enjoying the benefits of machine learning methods (e.g., adaptivity and improvement from experience).
Moreover, the \textsc{appl} framework explicitly allows for the possibility of adjusting these hyperparameters ``on-the-fly'', i.e., dynamically changing parameters at every time step to achieve robust navigation.
Finally, we also present a unifying vision for how our \textsc{appl} algorithms can be used in concert to enable a cycle-of-learning framework for continual system improvement through iterative interaction with simulation and end users.

We demonstrate the efficacy of the proposed algorithms in a series of experiments in which a small ground robot, accompanied by a human, must autonomously navigate in a number of complex, constrained environments.
Our results show that leveraging human interaction does indeed allow the robot to quickly adapt to each new environment, learning to overcome failures and other suboptimal behavior to the point where the human is no longer necessary.
% Moreover, we show that combining the proposed approaches in the unified framework that can also leverage unsupervised simulation results in a series of system iterates that exhibit increasing performance with each new deployment.
Moreover, we show that combining the proposed approaches in the unified framework that can also leverage unsupervised simulated training in a series of system iterations exhibits increasing performance with each new deployment. 

The remainder of this article is organized as follows. We begin by reviewing related work in terms of learning for navigation and learning from humans in Sec. \ref{sec::related}. As one of the two novel contributions of this article, we formalize the \emph{Adaptive Planner Parameter Learning} (\textsc{appl}) framework in Sec. \ref{sec::appl}, which is an overarching paradigm that encompasses all the following more specialized methods. 
% In Sec. \ref{sec::appld}, we briefly summarize the \textsc{appld} approach, \emph{Adaptive Planner Parameter Learning from Demonstration} \cite{xiao2020appld}. In Sec. \ref{sec::appli}, we provide details regarding \textsc{appli}, \emph{Adaptive Planner Parameter Learning from Interventions} \cite{wang2021appli}. In Sec. \ref{sec::apple}, we briefly summarize \textsc{apple}, \emph{Adaptive Planner Parameter Learning from Evaluative Feedback} \cite{wangapple}. In Sec. \ref{sec::applr}, we introduce \textsc{applr}, \emph{Adaptive Planner Parameter Learning from Reinforcement} \cite{xu2021applr}. 
Specifically, in Sec. \ref{sec::appld}, Sec. \ref{sec::appli}, Sec. \ref{sec::apple}, and Sec. \ref{sec::applr}, we present Adaptive Planner Parameter Learning from Demonstration \cite{xiao2020appld}, from Interventions \cite{wang2021appli}, from Evaluative Feedback \cite{wangapple}, and from Reinforcement \cite{xu2021applr}, respectively. 
Note that \textsc{appld} and \textsc{apple} were introduced in our previous Robotics and Automation Letters articles \cite{xiao2020appld} \cite{wangapple}, so we only briefly summarize these two methods. For \textsc{appli} and \textsc{applr}, which have only appeared in conference papers \cite{wang2021appli} \cite{xu2021applr}, we re-formalize these two methods under the overarching \textsc{appl} framework and present detailed descriptions. As the second novel contribution of this article, we introduce a cycle-of-learning scheme in Sec. \ref{sec::cycle}, which combines all four individual \textsc{appl} methods in the unified framework, leveraging demonstration, interventions, evaluative feedback, and reinforcement in different deployment scenarios with different human users, and achieves cyclic and continual improvement in navigation performance. The experiments pertaining to this cycle-of-learning scheme are also new in this article.
All experiment videos can be found at \url{https://www.cs.utexas.edu/~xiao/Research/APPL/APPL.html}
%%%%%%%%%%%%%%%%%%%%%%%%%%%%%%%%%%%%%%%%%%%%%%%%%%%%%%%%%%%%%%%%%%%%%%%%%%%%%%%%
\section{RELATED WORK}
\label{sec::related}

This section reviews state-of-the-art approaches that have applied machine learning to the problem of mobile robot navigation and related work in terms of using different human interaction modalities to assist machine learning. 

\subsection{Learning for Navigation}
Autonomous mobile robot navigation has been a topic of interest to the robotics community for decades \cite{quinlan1993elastic} \cite{fox1997dynamic}. 
For example, the DWA planner \cite{fox1997dynamic} samples feasible motion commands within a dynamic window, evaluates each sample using a forward prediction model, and selects the best sample based on a cost function that considers distance to obstacles, deviation from the global path, and progress toward the goal. 
While providing verifiable guarantees, these classical approaches still require extensive knowledge from robotics expert onsite during deployment, e.g., through in-situ parameter tuning, to adapt to different navigation scenarios \cite{zheng2017ros} \cite{xiao2017uav}. 
% They are also not good at enabling orthogonal capabilities to the original obstacle-avoidance-based navigation problem, e.g., social navigation and terrain-based navigation. 
For example, max linear and angular velocities and their sampling rates determine if the DWA planner can find good motion commands with limited onboard computation, while its different optimization weights affect the final navigation behavior. All experiments reported in this article use DWA as the underlying motion planning algorithm.  
Furthermore, those classical systems generally do not learn with increasing navigation experience \cite{liu2020lifelong}. 

With the recent advances in machine learning research, data-driven techniques have started being applied to the mobile robot navigation problem. Xiao, et al. \cite{xiao2020motion} presented a survey on using machine learning for motion control in mobile robot navigation: in addition to some works in the emerging social \cite{liang2020crowdsteer} and terrain-based \cite{xiao2021learning} navigation, most existing learning approaches for navigation adopt an end-to-end learning paradigm to address the classical collision-free navigation problem and are capable of generating navigational behaviors \cite{thrun1995approach} \cite{pfeiffer2017perception}. However, those end-to-end approaches still cannot outperform classical navigation methods or are not even compared to them. More importantly, end-to-end learning is extremely data-hungry, usually requiring millions of training data or training steps, and forgoes verifiable guarantees such as safety and explainability. On the other hand, other work which targeted a specific navigational component \cite{xiao2020agile} \cite{xiao2020toward} \cite{wangagile} has achieved superior performance when being compared to their classical counterparts. Therefore, it is promising to use machine learning at the subsystem or component level and to combine it with the structure of classical approaches \cite{xiao2020motion}. 

\textsc{appl} uses machine learning at the parameter level and devises extra learning components that interact with classical navigation systems. Therefore, \textsc{appl} inherits all the benefits of classical approaches, while enjoying the adaptivity and flexibility of learning methods.

\subsection{Learning from Humans}
Using interactions with humans is an effective approach to facilitate learning in general, e.g., Learning from Demonstration or Imitation Learning \cite{argall2009survey}.
Of particular interest to the robotics community is learning through interactions with non-expert users, which relaxes the requirement for expert roboticists. For mobile robot navigation, it is relatively easy for most non-expert users to provide a teleoperated demonstration \cite{siva2019robot}. If a robot can successfully navigate in most scenarios, it is only necessary to teach the robot where it makes mistakes. Learning from Intervention can therefore be applied only at those trouble-some situations \cite{goecks2019efficiently}. For non-expert users who are not able to take control of the robot, learning from evaluative feedback \cite{knox2013training} provides another interaction modality that teaches the robot with a simple scalar feedback value. Most aforementioned approaches of learning from humans have been proposed in the learning community and applied to test domains such as Atari games or simulations, and have yet been used to tackle physical mobile robot navigation problems. We posit that this gap can be caused by the requirement for extensive training data and training time, which is prohibitive for physical mobile robots navigating in the real world. 

\textsc{appl} utilizes all these human interaction modalities (demonstration, interventions, and evaluative feedback), and combines them with Reinforcement Learning (RL) in a Cycle-of-Learning scheme \cite{waytowich2018} for autonomous mobile robot navigation. To alleviate the heavy dependency on huge amounts of high quality training data from the real world, \textsc{appl} learns a \emph{parameter policy} that interacts with an underlying navigation system, in contrast to replacing it by learning an end-to-end motion policy. 

\subsection{Self-Supervised Learning} 
Researchers have also investigated self-supervised learning techniques for robotics. One such example is reinforcement learning from trial and error without access to expert or non-expert humans \cite{sutton2018reinforcement}. These self-supervised approaches work very well given access to a high-fidelity simulator \cite{chiang2019learning, xu2021machine} (e.g., in Atari games) and a well-defined reward function to specify the desired behavior to be learned \cite{knox2021reward}. Although such self-learning approaches remove the burden of having a human in the loop, they require an extensive amount of training data, which can be impractical to collect in the real world. Even with specific techniques to improve sample efficiency and allow real-world training, random exploration in the physical world is often very risky. For example, the \textsc{badgr} \cite{kahn2021badgr} system failed catastrophically (e.g., flipping over) during its exploration phase and required manual intervention to reset the robot. Another approach, \textsc{voila} \cite{karnan2021voila}, learns a visual navigation policy in the real world in a self-supervised manner with the guidance of the visual observation of another robot executing the same navigation task. But such methods don't easily generalize to other environments and other navigation tasks. Self-supervised model learning \cite{xiao2021learning} has also been investigated to learn kinodynamic models for real-world unstructured off-road terrain. But the self-supervised learning happens in an open-space, so that the probability of the robot colliding with any obstacles is minimized.

\textsc{appl} also leverages self-supervised learning in simulation using reinforcement learning, in addition to the non-expert human interactions collected from the wide spectrum of modalities in the real world. Such a Cycle-of-Learning scheme allows robots to continually improve their navigation behavior using both human interactions and self-supervision.

%%%%%%%%%%%%%%%%%%%%%%%%%%%%%%%%%%%%%%%%%%%%%%%%%%%%%%%%%%%%%%%%%%%%%%%%%%%%%%%%
\section{ADAPTIVE PLANNER PARAMETER LEARNING}
\label{sec::appl}

In this section, we formalize the \emph{Adaptive Planner Parameter Learning} (\textsc{appl}) problem, which serves as the foundation for all the following \textsc{appl} methods. 

\subsection{Underlying Navigation Planner}
\textsc{appl} assumes that a mobile robot has an underlying navigation planner $G: \mathcal{X} \times \Theta \rightarrow \mathcal{A}$, where $\mathcal{X}$ is the planner's state space (e.g., robot odometry, sensory inputs, navigation goal), $\Theta$ is the space of free parameters for $G$ (e.g., inflation radius, sampling rate, planner optimization coefficients), and $\mathcal{A}$ is the planner's action space (e.g., linear and angular velocity $v$ $\omega$). 
Using $G$ and a particular set of parameters $\theta$, the robot performs navigation by repeatedly estimating its state $x$ and applying action $a = G(x;\theta)= G_{\theta}(x)$.
Traditional learning methods for navigation replace classical navigation planner $G$ with a differentiable neural network, and use gradient descent to find thousands (or millions) of neural network weights. In contrast, \textsc{appl} works within the framework of classical planner $G$, but treats it as a black box, e.g., it does not need to be differentiable, and \textsc{appl} does not need to understand what each component of $\theta$ does.

\subsection{Meta-Environment}
\textsc{appl} works in the context of a \emph{meta-environment} $\mathcal{E}$ composed of both the underlying navigation world $\mathcal{W}$ (the physical, obstacle-occupied world) and the given classical planner $G$ with adjustable parameters $\theta \in \Theta$ and state input $x \in \mathcal{X}$. An \textsc{appl} agent interacts with this meta-environment $\mathcal{E}$ through a meta-state $s_t \in \mathcal{S}$, which includes both the planner's current state $x_t \in \mathcal{X}$ and previous parameters $\theta_{t-1} \in \Theta$, i.e., $\mathcal{S}=\mathcal{X}\times \Theta$. Instead of the raw action $a \in \mathcal{A}$, \textsc{appl}'s meta-action takes the form of $\theta \in \Theta$, which is the current parameters to be used by $G$. 

\subsection{Parameter Policy}
\begin{figure}
  \centering
  \includegraphics[width=\columnwidth]{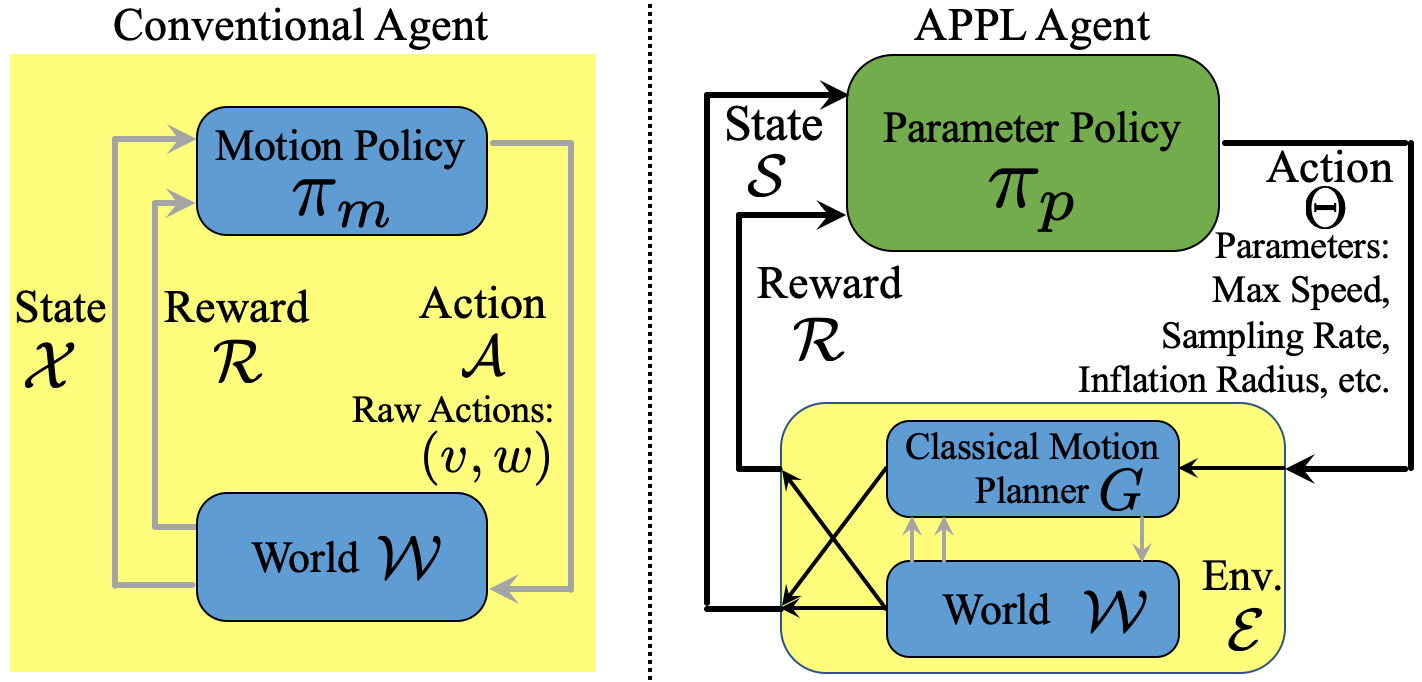}
  \caption{Contrast between Conventional Agent vs. \textsc{appl} Agent}
  \label{fig::appl}
\end{figure}

We formulate the \textsc{appl} problem as a meta Markov Decision Process (MDP) in this meta-environment, i.e., a tuple $(\mathcal{S}, \Theta, \mathcal{T}, \gamma, R)$. Note this meta-MDP differs from the conventional MDP defined by traditional learning-based motion planners, whose states are $x$ and actions are $a$ (Fig. \ref{fig::appl} left). Based on a meta-state $s_t$, an \textsc{appl} agent takes action $\theta_t$ so that the state advances to $s_{t+1}$ based on the transition function $s_{t+1} \sim \mathcal{T}(\cdot | s_t, \theta_t)$, and then receives a reward $r_t = R(s_t, \theta_t)$ (Fig. \ref{fig::appl} right). In general, the objective of an \textsc{appl} agent is to learn a \emph{parameter policy} $\pi: \mathcal{S} \rightarrow \Theta$ that can be used to select actions (as parameters $\theta$) that maximize the expected cumulative reward over time, 
\begin{equation}
    \max_\pi J^\pi = \mathbb{E}_{s_0, \theta_t \sim \pi(s_t), s_{t+1} \sim \mathcal{T}(s_t,\theta_t)}\bigg[\sum_{t=0}^\infty \gamma^t r_t\bigg].
    \label{eqn::reward}
\end{equation}
% \gw{This section is pretty good; the only comment I have is that the equation above is {\em not} the objective used for the ``parameter library" methods discussed before and I think that needs to be made clear in this section. I know you say ``in general" above, but I think it probably needs to be re-stated in the next section.}

Within the meta-environment $\mathcal{E}$, with the selected $\theta$, $G_\theta(x)$ produces navigation action $a$ to interact with the physical world $\mathcal{W}$. On top of this general form of parameter policy based on a general reward function, we impose additional structures on the policy and on the reward for each individual \textsc{appl} method. For example, \textsc{applr} adopts this general notion of parameter policy and optimizes a reward function; an \textsc{apple} policy maximizes the expected evaluative feedback from human users as an approximation of the underlying reward; \textsc{appli} and \textsc{appld} simplify the general parameter policy in terms of a pre-trained context predictor to select appropriate parameters from a parameter library. 

\subsection{Parameter Library}
A specific type of \textsc{appl} parameter policy $\pi: \mathcal{S} \rightarrow \Theta$ can be instantiated by imposing two intermediate mappings, i.e. a parameterized context predictor $B_\phi: \mathcal{S} \rightarrow \mathcal{C}$, and a one-to-one mapping $M: \mathcal{C} \rightarrow \Theta$, where $\mathcal{C}$ is the space of (finite) \emph{contexts} $c$ (a relatively cohesive region in the physical world $\mathcal{W})$. Therefore, $\theta = \pi(s)=M(B_\phi(s))$. Each predefined context is associated with a set of static parameters. These parameter sets comprise a \emph{parameter library} $\mathcal{L}$. \textsc{appld} and \textsc{appli} impose such structure on the general parameter policy, and selecting an appropriate parameter set which minimizes a Behavior Cloning loss approximates maximizing the underlying reward in Eqn. \ref{eqn::reward}.

The high-level algorithm of \textsc{appl} is shown in Alg. \ref{alg::appl}. The subroutine $\pi = LearnParameterPolicy(\mathcal{I}, \Theta, G) $ is instantiated differently in each \textsc{appl} method, as specified in Secs. \ref{sec::appld}-\ref{sec::applr}.

\begin{algorithm}
\caption{\textsc{appl}} \label{alg::appl}
\begin{algorithmic}[1]
    \STATE{// Training}
    \STATE{\textbf{Input:} human interaction $\mathcal{I}$, space of possible parameters $\Theta$, and navigation stack $G$.}
    \STATE $\pi = LearnParameterPolicy(\mathcal{I}, \Theta, G) $.
    % \STATE learn parameter policy $\pi: \mathcal{S} \rightarrow \Theta$ from $\mathcal{I}$ to fine-tune parameter $\theta \in \Theta$ of $G$.
    \STATE{// Deployment}
    \STATE \textbf{Input:} navigation stack $G$, parameter policy $\pi$.
    \FOR{$t=1:T$}
        \STATE construct meta-state $s_t$ from $x_t$ and $\theta_{t-1}$.
        \STATE $\theta_t = \pi(s_t)$.
        \STATE Navigate with $G_{\theta_t}(x_t)$.
    \ENDFOR
\end{algorithmic}
\end{algorithm}

%%%%%%%%%%%%%%%%%%%%%%%%%%%%%%%%%%%%%%%%%%%%%%%%%%%%%%%%%%%%%%%%%%%%%%%%%%%%%%%%
\section{APPL FROM DEMONSTRATION (\textsc{APPLD})}
\label{sec::appld}

\begin{figure}
  \centering
  \includegraphics[width=\columnwidth]{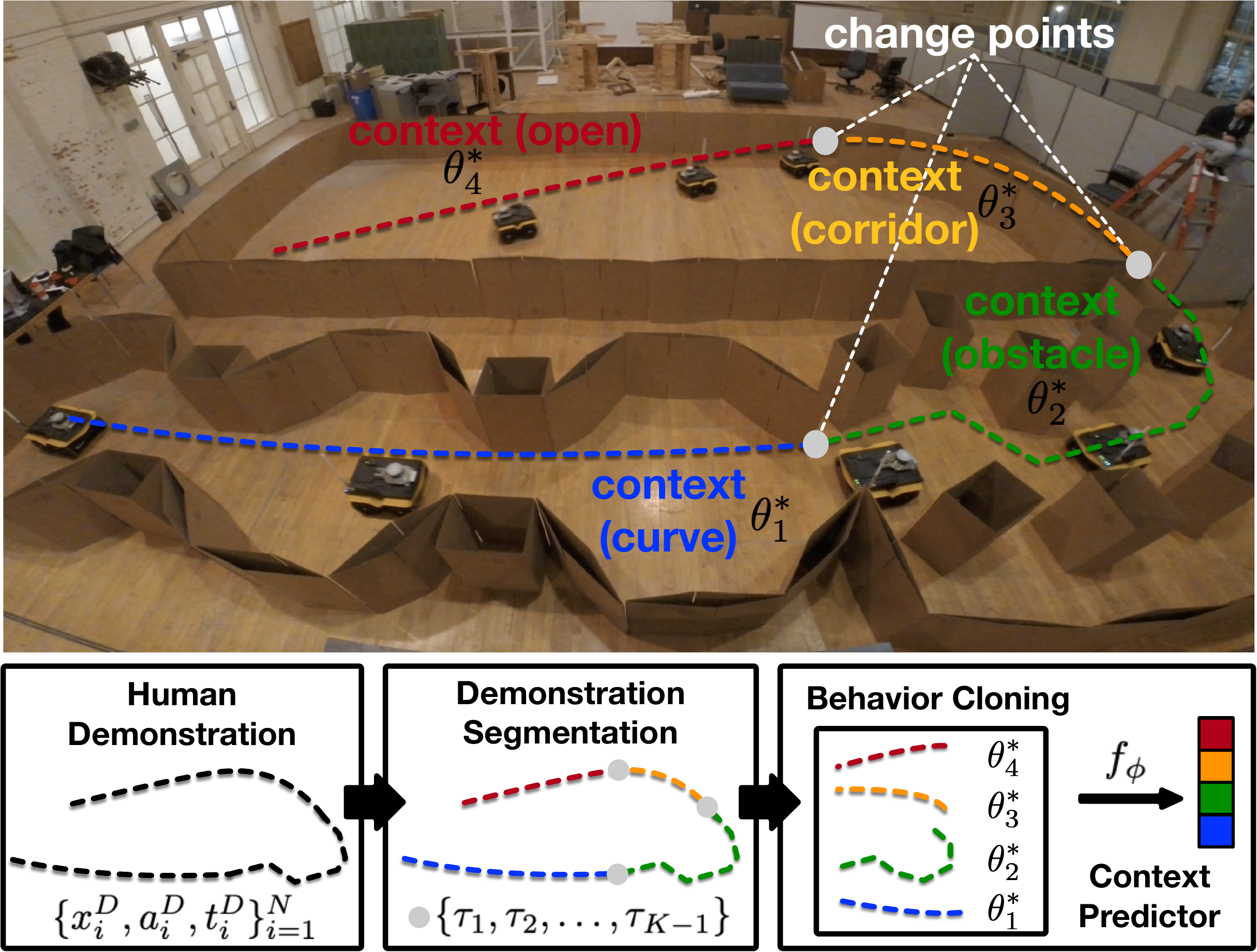}
  \caption{\textsc{appld}: a human demonstration is segmented into different contexts, for each of which, a set of parameters $\theta^*_k$ is learned via Behavior Cloning. During deployment, proper parameters are selected by an online context predictor.}
  \label{fig::jackal_ahg}
\end{figure}

In this section, we briefly summarize \emph{Adaptive Planner Parameter Learning from Demonstration} (\textsc{appld}) and re-formalize it based on the \textsc{appl} formulation in Sec. \ref{sec::appl}. \textsc{appld} \cite{xiao2020appld} utilizes a context predictor and a \emph{parameter library} learned from human demonstration. A teleoperated demonstration with planner state $x$ and demonstrated action $a$ is collected, which is first segmented into different \emph{contexts} using Bayesian change point detection. Within each context, we use behavior cloning to find a set of planner parameters $\theta$ to match the planner's output with the human demonstration (Fig. \ref{fig::jackal_ahg}). For full details and experiment results, please refer to our Robotics and Automation Letters (RAL) article \cite{xiao2020appld}. 

\subsection{Learning \textsc{appld} Policy}
A human demonstration of successful navigation is recorded as time series data $\mathcal{I}=\mathcal{D} = \{ x^D_i, a^D_i, t^D_i \}_{i=1}^N$, where $N$ is the length of the series, and $x^D_i$ and $a^D_i$ represent the planner state and demonstrated action at time $t^D_i$ (Line 2 Alg. \ref{alg::appl}).
As mentioned in Sec. \ref{sec::appl}, we impose additional structures on the \textsc{appld} policy, i.e. $\pi_D: \mathcal{S} \rightarrow \Theta$ is instantiated by a parameterized context predictor $B_\phi: \mathcal{S} \rightarrow \mathcal{C}$, and a one-to-one mapping $M: \mathcal{C} \rightarrow \Theta$ (Line 3 Alg. \ref{alg::appl}). Given $M$ and $B_\phi$, our system then performs navigation by selecting actions according to $G_\theta(x) = G_{M(B_\phi(x))}(x)$ (Lines 8-9 Alg. \ref{alg::appl}).

\subsubsection{Demonstration Segmentation}
\label{sec::appld_segmentation}
The key of learning $\pi_D$ is to construct the space of \emph{contexts} $\mathcal{C}$, each of which corresponds to a cohesive navigation region where a single set of parameters suffices. For each specific context, we apply black-box optimization to learn a parameter set. 
Generally speaking, any changepoint detection method can be used to solve this segmentation problem \cite{aminikhanghahi2017survey}. 
A changepoint detection algorithm $A_{\text{segment}}$ can automatically detect how many changepoints exist in $\mathcal{D}$ and where those changepoints are within the demonstration.
Denote the number of changepoints found by $A_{\text{segment}}$ as $K-1$ and the changepoints as $\tau_1, \tau_2, \dots, \tau_{K-1}$ with $\tau_0 = 0$ and $\tau_K = N+1$, the demonstration $\mathcal{D}$ is then segmented into $K$ pieces $\{\mathcal{D}_k=\{x^D_i, a^D_i, t^D_i \,|\, \tau_{k-1} \leq i < \tau_k\}\}_{k=1}^K$.

\subsubsection{Parameter Learning} 
\label{sec::appld_learning}
For each of the segmented contexts, $\mathcal{D}_k = \{x^D_i, a^D_i, t^D_i \,|\, \tau_{k-1} \leq i < \tau_k\}$, we employ {\em behavioral cloning} (\textsc{BC}) \cite{pomerleau1989alvinn} to learn a suitable set of parameters $\theta^*_k$.
More specifically, \textsc{BC} seeks $\theta^*_k$ which minimizes the difference between the demonstrated actions and the actions that $G_{\theta_k}$ would produce on $\{x^D_i\}$: 
\begin{equation}
    \begin{split}
    \theta^*_k &= \argmin_{\theta} \sum_{(x, a) \in \mathcal{D}_k} ||a - G_{\theta}(x))||_H,
    \end{split}
    \label{eqn::bc}
\end{equation}
where $||v||_H = v^THv$ is the induced norm by a diagonal matrix $H$ with positive real entries, which is used for weighting each dimension of the action. We solve Eqn. \ref{eqn::bc} with a black-box optimization method $A_{\text{black-box}}$. Having found each $\theta^*_k$, the \emph{parameter library} $\mathcal{L}$ is formed and the mapping $M$ is simply $M(k) = \theta^*_k$. This fully parallelizable optimization takes
approximately eight hours in our experiments on a single Dell XPS laptop
(Intel Core i9-9980HK) using 16 parallel threads, but this time could be significantly reduced with more computational resources and engineering effort.

\subsubsection{Online Context Prediction}
\label{sec::appld_context}
The context predictor $B_\phi$ is learned with a supervised dataset $\{x^D_i, c_i\}_{i=1}^N$, where $c_i = k$ if $x^D_i \in \mathcal{D}_k$. To classify which segment $x^D_i$ comes from, we learn a parameterized function $f_\phi(x)$ via supervised learning:  
\begin{equation}
    \phi^* = \argmax_\phi \sum_{i=1}^N \log \frac{\exp\big(f_\phi(x^D_i)[c_i]\big)}{\sum_{c=1}^K \exp{\big(f_\phi(x^D_i)[c]\big)}}.
\end{equation}
Our context predictor $B_\phi$ is then defined as: 
\begin{equation}
    B_\phi(x_t) = \mode \Big\{\argmax_{c} f_\phi(x_i)[c],\,\, t-p < i \leq t \Big\}.
    \label{eqn:mode_switching}
\end{equation}
In other words, $B_\phi$ acts as a mode filter on the context predicted by $f_\phi$ over a sliding window of length $p$.

The $LearningParameterPolicy$ subroutine in Alg. \ref{alg::appl} for \textsc{appld} is shown in Alg. \ref{alg::appld}, where the above three stages are applied sequentially to learn a \emph{parameter library} $\mathcal{L} = \{\theta^*_k\}_{k=1}^K$ (hence the mapping $M$) and a \emph{context predictor} $B_\phi$, both of which constitute the \textsc{appld} policy $\pi_D$. During deployment, Eqn. \ref{eqn:mode_switching} is applied online to pick the right set of parameters for $G$. $\pi_D$, as a special case of $\pi$, does not consider $\theta_{t-1}$ as part of $s_t$ (only $x_t$, line 7 Alg. \ref{alg::appl}), but consults a history of $p$ steps of $x_t$ for smoothness. 

\begin{algorithm}
\caption{$LearnParameterPolicy$ (\textsc{appld})} \label{alg::appld}
\begin{algorithmic}[1]
    \STATE{\textbf{Input:} $\mathcal{I} = \mathcal{D} = \{ x^D_i, a^D_i, t^D_i \}_{i=1}^N$, $\Theta$, $G$.}
    \STATE Call $A_{\text{segment}}$ on $\mathcal{D}$ to detect changepoints $\tau_1, \dots, \tau_{K-1}$ with $\tau_0=0$ and $\tau_K=N+1$.
    \STATE Segment $D$ into $\{\mathcal{D}_k=\{x^D_i, a^D_i, t^D_i \,|\, \tau_{k-1} \leq i < \tau_k\}\}_{k=1}^K$.
    \STATE Train a classifier $f_\phi$ on $\{x^D_i, c_i\}_{i=1}^N$, where $c_i = k$ if $x^D_i \in \mathcal{D}_k$.
    \FOR{$k=1:K$}
        \STATE Call $A_{\text{black-box}}$ with objective defined in Eqn.  \ref{eqn::bc} on $\mathcal{D}_k$ to find parameters $\theta^*_k$ for context $k$.
    \ENDFOR
    \STATE Form the map $M(k) = \theta^*_k$, $\forall 1\leq k\leq K$ and context predictor $B_\phi(x)$.
\end{algorithmic}
\end{algorithm}

\subsection{Experiments}
\label{sec::appld_experiments}
We implement \textsc{appld} on a physical ClearPath Jackal robot to experimentally validate that using the learned parameter library and context predictor from a teleoperated demonstration can achieve better navigation performance in complex environments compared to that obtained by the underlying navigation system using {\em (a)} its default parameters from the robot platform manufacturer, and {\em (b)} parameters we found using behavior cloning but without context. The four-wheeled, differential-drive, unmanned ground vehicle with a top speed of 2.0m/s is tasked to move through a custom-built maze as quickly as possible (Fig. \ref{fig::jackal_ahg}). A Velodyne LiDAR provides 3D point cloud data, which is transformed into 2D laser scan for 2D navigation. 
The Jackal runs Robot Operating System (\textsc{ros}) onboard, and \textsc{appld} is applied to the local planner, \textsc{dwa} \cite{fox1997dynamic}, in the commonly-used \texttt{move\textunderscore base} navigation stack. Other parts of the navigation stack, e.g. global planning with Dijkstra's algorithm, remain intact.

We use \textsc{champ} as $A_{\text{segment}}$ (Line 2 Alg. \ref{alg::appld}), a state-of-the-art Bayesian segmentation algorithm \cite{niekum2015online}. We find each $\theta_k^*$ using \textsc{cma-es} \cite{hansen2003reducing} as our black-box optimizer (Line 6 Alg. \ref{alg::appld}). $\theta$ includes \textsc{dwa}'s \emph{max\_vel\_x} (v), \emph{max\_vel\_theta} (w), \emph{vx\_samples} (s), \emph{vtheta\_samples} (t), \emph{occdist\_scale} (o), \emph{pdist\_scale} (p), \emph{gdist\_scale} (g) and costmap's \emph{inflation\_radius} (i).

In our experiments, \textsc{appld} achieves superior performance in terms of fastest traversal time compared to the default parameters, parameters learned  without context, and even the demonstrator. For all the experiments in this paper, we use traversal time as the comparison metric since most suboptimal navigation behavior cause stop-and-go motions, induce recovery behaviors, cause the robot to get stuck, or collide with obstacles (termination) - each of which will result in a higher traversal time. For full details about the experimental setup and results, along with experimental results on a different robot with a different navigation system in a different environment, please refer to our RAL article \cite{xiao2020appld}. 

\section{APPL FROM INTERVENTIONS (\textsc{APPLI})}
\label{sec::appli}
\begin{figure}
  \centering
  \includegraphics[width=\columnwidth]{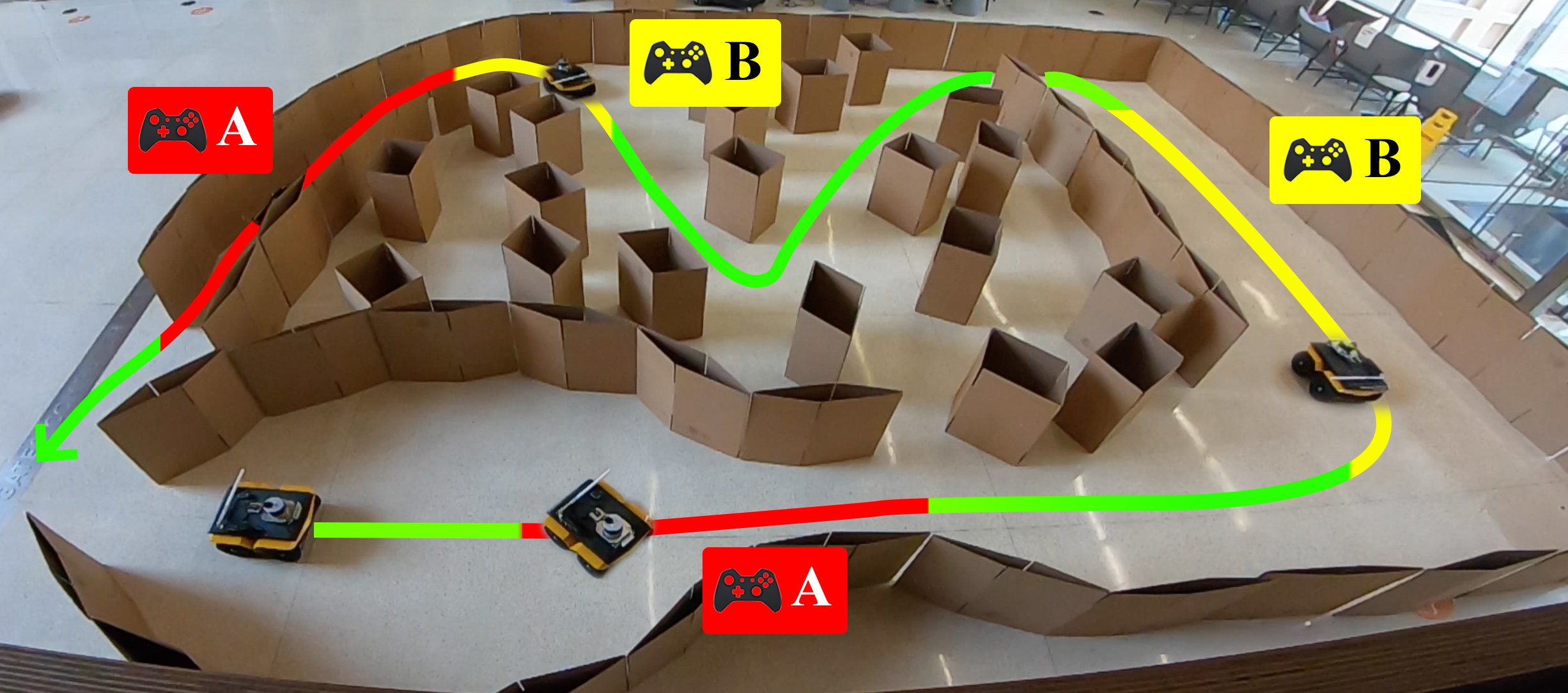}
  \caption{While classical navigation systems perform well in most places (green), they may fail (red) or suffer from suboptimal behavior (yellow) in others. \textsc{appli} learns from human interventions in these two scenarios (Type A and Type B).}
  \label{fig::appli}
\end{figure}

\textsc{appld} assumes human demonstration is a good approximation for optimal navigation behavior and default parameters do not work well in most places, which are not always the case. 
% Default parameters of classical navigation planners work well in most cases, while may fail or achieve suboptimal behavior only in certain places. 
% Furthermore, a full human demonstration is not always optimal. \gw{Transition from APPLD to here can be smoother.} 
In \emph{Adaptive Planner Parameter Learning from Interventions} (\textsc{appli}) \cite{wang2021appli}, instead of providing a full demonstration of the entire navigation task (such as \textsc{appld}), non-expert users can easily identify failure or suboptimal cases by watching and then provide a few teleoperated \emph{interventions} to correct the failure or suboptimal behaviors (Fig. \ref{fig::appli}). \textsc{appli} utilizes this human interaction modality and learns sets of planner parameters specifically for the scenarios where failure and suboptimal behaviors take place, to create a \emph{parameter library} including the default parameters. During deployment, \textsc{appli} applies those learned parameters, only when it is confident that they will benefit the current navigation based on a confidence measure of the context predictor, to the underlying navigation system in those troublesome places, while maintaining good performance in others by switching back to the default parameters. This confidence measure also enables \textsc{appli} to generalize well to unseen environments.

\subsection{Learning \textsc{appli} Policy}
\label{sec::appli_learning}
\textsc{appli} assumes a default parameter set $\bar{\theta}$ is tuned by a human designer trying to achieve good performance in most environments. However, being good at everything often means being great at nothing: $\bar{\theta}$ usually exhibits suboptimal performance in some situations and may even fail (is unable to find feasible motions, or crashes into obstacles) in particularly challenging ones.

\subsubsection{Interventions as Contexts}
To mitigate this problem, a human can supervise the navigation system's performance at state $x$ by observing its action $a$ and judging whether she should intervene. 
Here, we consider two types of interventions. A Type A intervention is one in which the system performs so poorly that the human {\em must} intervene (e.g. imminent collision or a signal for help). A Type B intervention is one in which a human {\em might} intervene in order to improve otherwise suboptimal performance (e.g., driving too slowly in an open space). 
For the $i^\text{th}$ intervention, we assume that the human resets the robot to the position where the failure or suboptimal behavior first occurred and then gives a short teleoperated intervention $I_i=\{x_t, a_t\}_{t=1}^{T_i}$ of length $T_i$, where $x_{1:T_i}$ is the trajectory starting from the reset state induced by intervention actions $a_{1:T_i}$. As this short demonstration naturally shows a cohesive navigation behavior in a specific segment of the environment (open space, narrow corridor, etc), it automatically forms a \textit{context} $c_i \in \mathcal{C}$. Using human interaction composed of $N$ interventions $\mathcal{I} = I_{1:N}$ instead of the full demonstration sequence $\mathcal{D} = \{ x^D_i, a^D_i, t^D_i \}_{i=1}^N$ (Sec. \ref{sec::appld}), \textsc{appli} finds the mapping $M: \mathcal{C} \rightarrow \Theta$ that determines the parameter set $\theta_i$ for each intervention context $c_i$, and the parameterized context predictor $B_\phi: \mathcal{X} \rightarrow \mathcal{C}$ that determines to which context (if any) the current state $x$ belongs. 
After collecting a set of $N$ interventions $I_{1:N}$, for each $I_i$, we learn a set of navigation parameters $\theta_i$ that can best imitate the demonstrated correction with the same Behavior Cloning loss (Eqn. \ref{eqn::bc}) with a black-box optimizer, such as \textsc{cma-es} \cite{hansen2003reducing}. After identifying parameters in each context, the mapping $M$ is simply $M(i) = \theta_i$. 

\subsubsection{Confidence-Based Context Prediction}
\label{sec::appli_context}
In contrast to \textsc{appld}, \textsc{appli} does not learn parameters for places where the original navigation performance is good. 
Therefore, it determines if the current state $x_t$ falls into any one of the collected intervention contexts $c_i$. If so, \textsc{appli} directs the robot to use the parameter set $\theta_i$ to avoid making the same mistake as before. If not, \textsc{appli} directs the robot to use the default parameters $\bar{\theta}$, as they are optimized for most cases and are expected to generalize better than any parameter set learned for a specific scenario. In our system, the determination above is made using a new context predictor, $B_\phi$, with confidence measure. To train this predictor, we build a similar dataset as in \textsc{appld}, $\{\{x_t,c_i\}_{t=1}^{T_i}\}_{i=1}^N$, and train an intermediate classifier $f_\phi(x)$ with parameter set $\phi$ using the Evidential Deep Learning method (EDL) \cite{sensoy2018evidential}. A feature of EDL is that it supplies both a predicted label and a confidence measure, $u_i \in (0, 1]$, i.e.,
\begin{equation}
    f_\phi(x_i) = (c_i, u_i).
\end{equation}
After training $f_\phi$ and during deployment, we build a confidence-based classifier $g_\phi$ as 
\begin{equation}
    g_\phi(x_i) = c_i \mathbbm{1}(u_i \ge \epsilon_u),
\end{equation}
where $\epsilon_u$ is a threshold on confidence and $\mathbbm{1}$ is the indicator function. For state $x_i$, $g_\phi$ determines its context from $N+1$ contexts ($N$ intervention contexts and one default context). If $u_i \ge \epsilon_u$, it suggests the classifier $f_\phi$ is confident and $g_\phi$ predicts $c_i$. Otherwise, when $f_\phi$ is unsure about its prediction, $c_i \mathbbm{1}\{u_i \ge \epsilon_u \} = 0$. In this case, $g_\phi$ believes the current state $x_i$ is not similar to any intervention context and instead classifies $x_i$ as the default context. For this default context labeled as $c_i=0$, navigation utilizes the default navigation parameters $\bar{\theta}$ (i.e., we set $M(0) = \bar{\theta}$). 
The new confidence-based context predictor $B_\phi$ is then defined as:
\begin{equation}
    % B_\phi(x_t) = \text{mode}( \left\{ g_\phi(x_i) \right\}_{i = t - w + 1}^t).
    B_\phi(x_t) = \mode \Big\{g_\phi(x_i),\,\, t-p < i \leq t \Big\}.
    \label{eq:context_predictor}
\end{equation}
Again, $B_\phi$ acts as a mode filter and chooses the context $c_t$ that the majority agrees with over the past $p$ time steps.  

The $LearningParameterPolicy$ subroutine in Alg. \ref{alg::appl} for \textsc{appli} is shown in Alg. \ref{alg::appli}. \textsc{appli} learns a confidence-based classifier (Line 2) and navigation parameters $\theta_{1:N}$ for each context (Lines 3-5). During deployment, we use $\theta_t = \pi_I(x_t) = M(B_\phi(x_t))$ to select the parameters for the navigation system at time $t$. Similar to $\pi_D$, $\pi_I$ does not consider $\theta_{t-1}$ as part of $s_t$ (only $x_t$, line 7 Alg. \ref{alg::appl}), but consults a history of $p$ steps of $x_t$ for smoothness. 
% does not consider $\theta_{t-1}$ in addition to $x_t$, and consults a history of $p$ steps for smoothness. 

\begin{algorithm}
\caption{$LearningParameterPolicy$ (\textsc{appli})} \label{alg::appli}
\begin{algorithmic}[1]
\STATE{\textbf{Input:} $\mathcal{I} = I_{1:N} = \{\{x_t,a_t\}_{t=1}^{T_i}\}_{i=1}^N$, $\Theta$, $G$.}
\STATE Train a confidence-based classifier $f_\phi$ on $\{\{x_t,c_i\}_{t=1}^{T_i}\}_{i=1}^N$.\\
\FOR{$i = 1, \dots, N$}
    \STATE Find parameter $\theta_i$ for context $i$ using Eqn. \ref{eqn::bc} on $I_i$.
\ENDFOR
\STATE Form the map $M(i)=\theta_i$, $\forall 1\leq i\leq N$, and confidence-based context predictor $B_\phi(x)$.
\end{algorithmic}
\end{algorithm}

\subsection{Experiments}
\label{sec::appli_experiments}
In our experiments, we aim to show that \textsc{appli} can improve navigation performance by learning from only a few interventions and, with the confidence measurement, that the overall system can generalize well to unseen environments. 
We apply \textsc{appli} on the same ClearPath Jackal ground robot with the same setup as in \textsc{appld} (Section \ref{sec::appld_experiments}) in a physical obstacle course. Navigation performance learned through \textsc{appli} is then tested both in the same training environment, and also in another unseen physical test course, which is qualitatively similar to the training environment (i.e., similar contexts were created in a different ordering). Furthermore, to investigate generalizability, we test the learned systems on a benchmark suite of 300 unseen simulated navigation environments \cite{perille2020benchmarking}. 

During data collection, one of the authors (the {\em intervener}) follows the robot through the test course and intervenes when necessary, reporting if the intervention is to drive the robot out of a failure case (Type A) or to correct a suboptimal behavior (Type B). The four interventions are shown in Fig. \ref{fig::appli}: before the two Type A interventions (shown in red), the default system (\textsc{dwa} with $\bar{\theta}$) fails to plan feasible motions and starts recovery behaviors (rotates in place and moves backward); before the two Type B interventions (shown in yellow), the robot drives unnecessarily slowly in a relatively open space and enters the narrow corridor with unsmooth motions. For every intervention, the intervener stops the robot, drives it back to where they deem the failure or suboptimal behavior to have begun, and then provides recorded teleoperation $I$ that avoids the problematic behavior. It takes less than an hour to learn a set of parameters for each intervention. Training the EDL-based context predictor takes only a few minutes using the same computational infrastructure as specified in Section \ref{sec::appld_segmentation}. To compare the performance learned from interventions and learned from a full demonstration, we also collect extra demonstrations for those places where the default planner already works well (shown in green in Fig. \ref{fig::appli}).

\subsubsection{Physical Experiments}
After training, we deploy $\pi_I$ with learned mapping $M$ and context predictor $B_\phi$ on the \texttt{move\textunderscore base} navigation stack $G$ with an empirically chosen threshold value $\epsilon_u=0.8$. 

\begin{figure}
  \centering
  \includegraphics[width=\columnwidth]{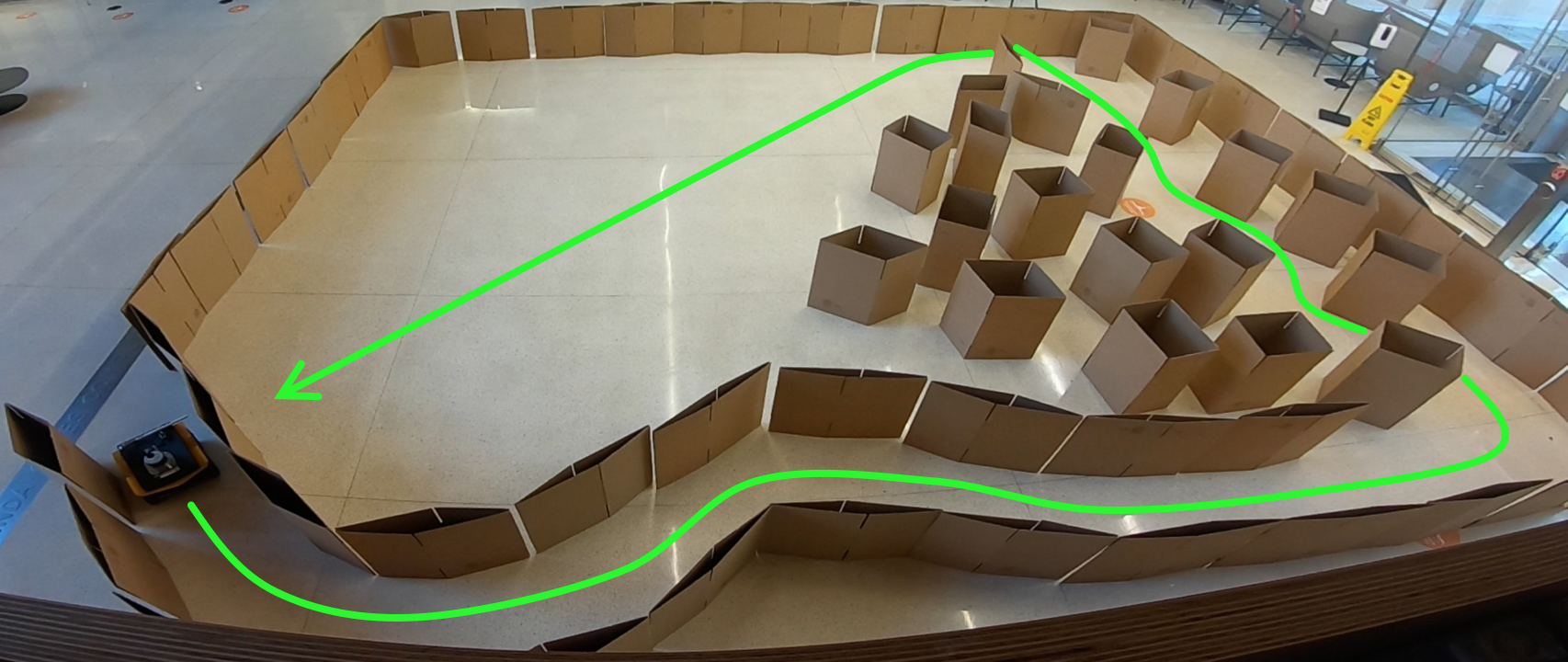}
  \caption{\textsc{appli} Running in an Unseen Physical Environment}
  \label{fig::unseen}
\end{figure}

We first deploy \textsc{appli} in the same training physical environment (Fig. \ref{fig::appli}). We compare the performance of the \textsc{dwa} planner with default parameters, \textsc{appli} learned only with Type A interventions, \textsc{appli} learned with Type A \emph{and} Type B interventions, and \textsc{appli} learned with a \emph{full demonstration} (which is basically \textsc{appld} enhanced by manual context segmentation and the confidence measure).
The motivation for the variation of \textsc{appli} learned only with Type A interventions is to study the effect of an unfocused or inexperienced human intervener. In this case, the human would still conduct all Type A interventions, as those mistakes are severe and easy to identify---some robots may even actively ask for help (e.g. by starting recovery behaviors). However, the human may fail to conduct Type B interventions as she is not paying attention, or isn't equipped with the knowledge to identify suboptimal behaviors.
For each method, we run five trials and report the mean and standard deviation of the traversal time in Tab. \ref{tab::training_results} 1st row. If the robot gets stuck, we introduce a penalty value of 200 seconds. We also deploy the same sets of variants in an unseen physical environments (Fig. \ref{fig::unseen} and Tab. \ref{tab::training_results} 2nd row). 

\begin{table*}
\centering
\caption{\textsc{appli} Traversal Time}
\begin{tabular}{ccccc}
\toprule
& Default & Type A & Type A+B & Full Demo \\ 
\midrule
Training Environment& 134.0$\pm$60.6s & 77.4$\pm$2.8s  & 70.6$\pm$3.2s & 78.0$\pm$2.7s \\
Unseen Environment& 109.2$\pm$50.8s & 71$\pm$0.7s & 59$\pm$0.7s & 62.0$\pm$2.0s\\
\bottomrule
\end{tabular}
\label{tab::training_results}
\end{table*}

% \begin{table}
% \centering
% \caption{Traversal Time in Unseen Environment}
% \begin{tabular}{cccc}
% \toprule
% Default & Type A & Type A+B & Full Demo\\ 
% \midrule
% 109.2$\pm$50.8s & 71$\pm$0.7s & 59$\pm$0.7s & 62.0$\pm$2.0s\\
% \bottomrule
% \end{tabular}
% \label{tab::unseen_results}
% \end{table}

For both the training and unseen environments, Type A interventions alone significantly improve upon the default parameters ($p = 0.016$), by correcting all recovery behaviors such as rotating in place or driving backwards, and eliminating all failure cases. Adding Type B interventions further reduces traversal time ($p = 5 \times 10^{-8}$), since the robot learns to speed up in relatively open spaces and to execute smooth motion when the tightness of the surrounding obstacles changes. All the interventions are able to improve navigation in both training and unseen environments ($p_\text{env.} = 0.11$), suggesting \textsc{appli}'s generalizability. Surprisingly, in both environments, \textsc{appli} learned from only Type A and Type B interventions can even outperform \textsc{appli} learned from an entire demonstration ($p = 1.5 \times 10^{-4}$). One possible reason for this better performance from fewer human interactions is the additional human demonstrations may be suboptimal, especially since they are collected in places where the default navigation system was already deemed to have performed well. 
For example, in the full demonstration, we find the human intervener is more conservative than the default navigation system and drives slowly in some places. Hence, learning from these suboptimal behaviors introduces suboptimal parameters and consequently worse performance in contexts similar to that intervention. 

\subsubsection{Simulated Experiments}
\begin{figure*}
  \centering
  \includegraphics[width=1\columnwidth]{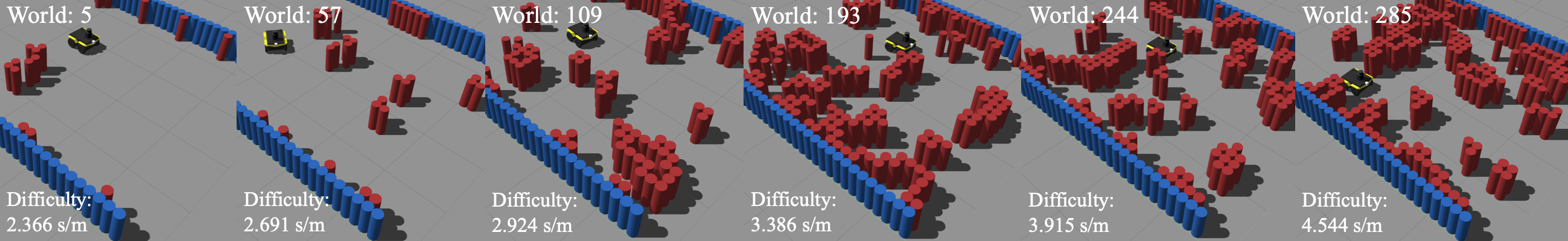}
  \caption{Example Navigation Environments in the \textsc{barn} Dataset}
  \label{fig::simulated_envs}
\end{figure*}

\iffalse
\begin{table*}[t]
\small
\vspace{0.25cm}
\centering
\caption{Percentage of Simulation Environments where Method 1 is Significantly Worse than Method 2 in terms of Traversal Time\\
\footnotesize (Methods are listed in order of increasing performances. Results mentioned in experiment analysis are bold for better identification)}
\begin{tabular}{c l c c c c c c c}
\toprule
  & & \multicolumn{7}{c}{\footnotesize Method 2} \\
  & & \footnotesize \textsc{appli} (A) & \footnotesize \textsc{dwa} & \footnotesize \textsc{appli} (A+c) & \footnotesize \textsc{appli} (A+B+D+c) & \footnotesize \textsc{appli} (A+B+D) & \footnotesize \textsc{appli} (A+B+c) & \footnotesize \textsc{appli} (A+B)\\ 
\midrule
\multirow{7}{*}{\shortstack{\footnotesize Method\\1}} & \footnotesize \textsc{appli} (A) & 0   & 50    & \textbf{53}    & 62    & 63    & 68    & 66 \\
& \footnotesize \textsc{dwa}           & 10   & 0    & 6    & \textbf{33}    & 40    & \textbf{44}    & 47 \\
& \footnotesize \textsc{appli} (A+c)   & 6   & 4    & 0    & 31    & 37    & 45    & 45 \\
& \footnotesize \textsc{appli} (A+B+D+c) & 5   & 7    & 11    & 0    & 25    & \textbf{31}    & 33 \\
& \footnotesize \textsc{appli} (A+B+D) & 5   & 7    & 7    & 10    & 0    & \textbf{21}    & 21 \\
& \footnotesize \textsc{appli} (A+B+c) & 3   & 3    & 4    & 3    & 5    & 0   & 9 \\
& \footnotesize \textsc{appli} (A+B)   & 2   & 5    & 5    & 6    & 4    & 6   & 0 \\
\bottomrule
\end{tabular}
\label{tab::similation_results}
\end{table*}
\fi
To further test \textsc{appli}'s generalizability to unseen environments, we test our method and compare it with two baselines on the Benchmark for Autonomous Robot Navigation (\textsc{barn}) dataset \cite{perille2020benchmarking}. The benchmark dataset consists of 300 simulated navigation environments randomly generated using Cellular Automata, ranging from easy ones with a lot of open spaces to challenging ones where the robot needs to get through dense obstacles (see examples in Fig. \ref{fig::simulated_envs}). 
% Example environments with increasing difficulty levels are shown in Fig. \ref{fig::simulated_envs}. 
Using the same training data collected from the physical environment shown in Fig. \ref{fig::appli}, we test the following seven variants: 
(1) \textsc{appli} (A+B+c): \textsc{appli} learned from \emph{Type A and B} inventions \emph{with} confidence measure, 
(2) \textsc{appli} (A+B): \textsc{appli} learned from \emph{Type A and B} inventions \emph{without} confidence measure, 
(3) \textsc{appli} (A+c): \textsc{appli} learned from only \emph{Type A} interventions \emph{with} confidence measure, 
(4) \textsc{appli} (A): \textsc{appli} learned from only \emph{Type A} interventions \emph{without} confidence measure,
(5) \textsc{appli} (A+B+D+c): \textsc{appli}  learned from \emph{full demonstration} \emph{with} confidence measure,
(6) \textsc{appli} (A+B+D): \textsc{appli}  learned from \emph{full demonstration} \emph{without} confidence measure, 
(7) the \textsc{dwa} planner with default parameters.

\begin{figure}
  \centering
  \includegraphics[width=\columnwidth]{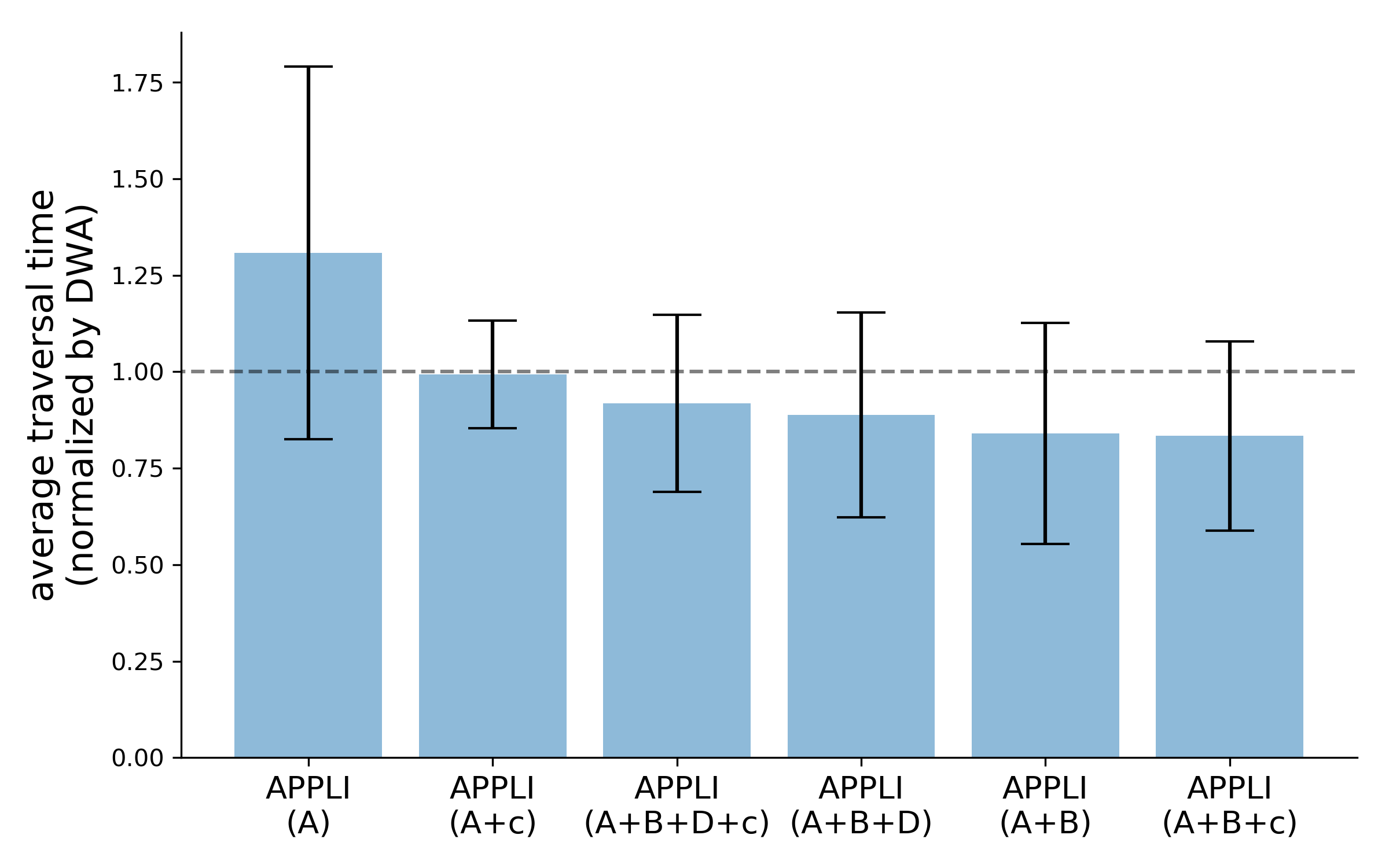}
  \caption{Normalized Performance in 300 Simulation Environments from 12 Runs Each. Error bars: standard deviation.}
  \label{fig::appli_simulated}
\end{figure}

\begin{table*}
\small
\vspace{0.25cm}
\centering
\caption{Percentage of Simulation Environments that Method 1 is Significantly Worse than Method 2 in Terms of Traversal Time\\
\scriptsize (Methods are listed in order of increasing performances. Results mentioned in experiment analysis are bold for better identification)}
\begin{tabular}{c c c c c c c c c}
\toprule
  & & \multicolumn{7}{c}{Method 2} \\
  & & (A) & \textsc{dwa} & (A+c) & (A+B+D+c) & (A+B+D) & (A+B+c) & (A+B)\\ 
  & & \rom{1} & \rom{2} & \rom{3} & \rom{4} & \rom{5} & \rom{6}& \rom{7}\\ 
\midrule
\parbox[t]{2mm}{\multirow{7}{*}{\rotatebox[origin=c]{90}{Method 1}}} & \rom{1} & 0   & 50    & \textbf{53}    & 62    & 63    & 68    & 66 \\
& \rom{2}           & 10   & 0    & 6    & \textbf{33}    & 40    & \textbf{44}    & 47 \\
& \rom{3}   & 6   & 4    & 0    & 31    & 37    & 45    & 45 \\
& \rom{4} & 5   & 7    & 11    & 0    & 25    & \textbf{31}    & 33 \\
& \rom{5} & 5   & 7    & 7    & 10    & 0    & \textbf{21}    & 21 \\
& \rom{6} & 3   & 3    & 4    & 3    & 5    & 0   & 9 \\
& \rom{7}   & 2   & 5    & 5    & 6    & 4    & 6   & 0 \\
\bottomrule
\end{tabular}
\label{tab::similation_results}
\end{table*}

Testing these variations aims at studying the effect of learning from different modes of interventions caused by different degrees of human attention and experience levels, i.e., imperative interventions (A), optional interventions (A + B), and a full demonstration (A + B + D). They also provide an ablation study for the confidence measure in the EDL context classifier $f_\phi$: when deployed without the confidence measure, the robot has to choose among the parameters learned from interventions and never uses the default parameters. 

For each method in each simulation environment, we measure the traversal time for 12 different runs, resulting in 25200 total navigation trials. 
% The average traversal time for each method in all simulation environments are shown in order of increasing performance in Tab. \ref{tab::simulation_mean}. 
The average traversal time for each method in all simulation environments is normalized by DWA and is shown in order of increasing performance in Fig. \ref{fig::appli_simulated}.
We conduct a pair-wise t-test for all methods in order to compute the percentage of environments in which one method (denoted as Method 1) is significantly worse ($p<0.05$) than another (denoted as Method 2). For better illustration, we also reorder the method by performance and show the pairwise comparisons in Tab. \ref{tab::similation_results}.

% \textsc{appli} (A+B+c) and \textsc{appli} (A+B+D+c) outperform \textsc{dwa}: they are significantly better in 44\% and 33\% of environments respectively and significantly worse in only 3\% and 7\% of environments than \textsc{dwa}.
% However, \textsc{appli} (A+c) is only significantly better than \textsc{dwa} 6\% of the time, which suggests that even though type B inventions only correct suboptimal performances, they are crucial for performance improvement. 

\textsc{appli} (A+B+c) and \textsc{appli} (A+B+D+c) outperform \textsc{dwa}: their average traversal time is shorter than that of \textsc{dwa} by 8\% and 17\% respectively, and they are significantly better in 44\% and 33\% of environments respectively and significantly worse in only 3\% and 7\% of environments than \textsc{dwa}.
However, for \textsc{appli} (A+c), its traversal time is only 1\% better than \textsc{dwa} (significantly better 6\% of the time), which suggests that even though type B inventions only correct suboptimal performances, they are crucial for performance improvement. 

In terms of the effect of confidence, 
\textsc{appli} (A) only selects parameters learned from 2 Type A inventions and never uses the default parameters even when they are more appropriate. % Removing confidence greatly harms its performance, making it significantly worse than \textsc{appli} (A+c) in 53\% of environments. 
Removing confidence greatly harms its performance, making its traversal time even longer than \textsc{dwa} by 31\% (significantly worse in 53\% of environments). 
However, \textsc{appli} (A+B+c) and \textsc{appli} (A+B+D+c), which use more interventions or even the full demonstration to train the parameter mapping $M$ and context predictor $B_\phi$, are more confident about their predictions most of the time. As a result, removing confidence in the context predictor doesn't result in a significant difference.

% Lastly, a counteractive, but similar result as in the physical experiments is that compared with \textsc{appli} (A+B+D+c) which uses the full demonstration, \textsc{appli} (A+B+c) learned from only Type A and B interventions achieves superior performance by being significantly better than \textsc{appli} (A+B+D+c) and \textsc{appli} (A+B+D) in 31\% and 21\% of the environments,  respectively. 
Lastly, a counterintuitive, but similar result as in the physical experiments is that 
% compared with \textsc{appli} (A+B+D+c) which uses the full demonstration, 
\textsc{appli} (A+B+c) learned from only Type A and B interventions achieves shorter traversal time than \textsc{appli} (A+B+D+c) and \textsc{appli} (A+B+D) by 9.2\% and 6.1\% (significantly better in 31\% and 21\% of the environments), respectively.
Similar to the discussions about physical experiments, unnecessary human demonstrations are most likely suboptimal. In this sense, \textsc{appli} not only reduces the required human interactions from a full demonstration to only a few interventions, but also reduces the chance of performance degradation caused by suboptimal demonstrations.

%%%%%%%%%%%%%%%%%%%%%%%%%%%%%%%%%%%%%%%%%%%%%%%%%%%%%%%%%%%%%%%%%%%%%%%%%%%%%%%%
\section{APPL FROM EVALUATIVE FEEDBACK (\textsc{APPLE})}
\label{sec::apple}
\begin{figure}[ht]
  \centering
  \includegraphics[width=\columnwidth]{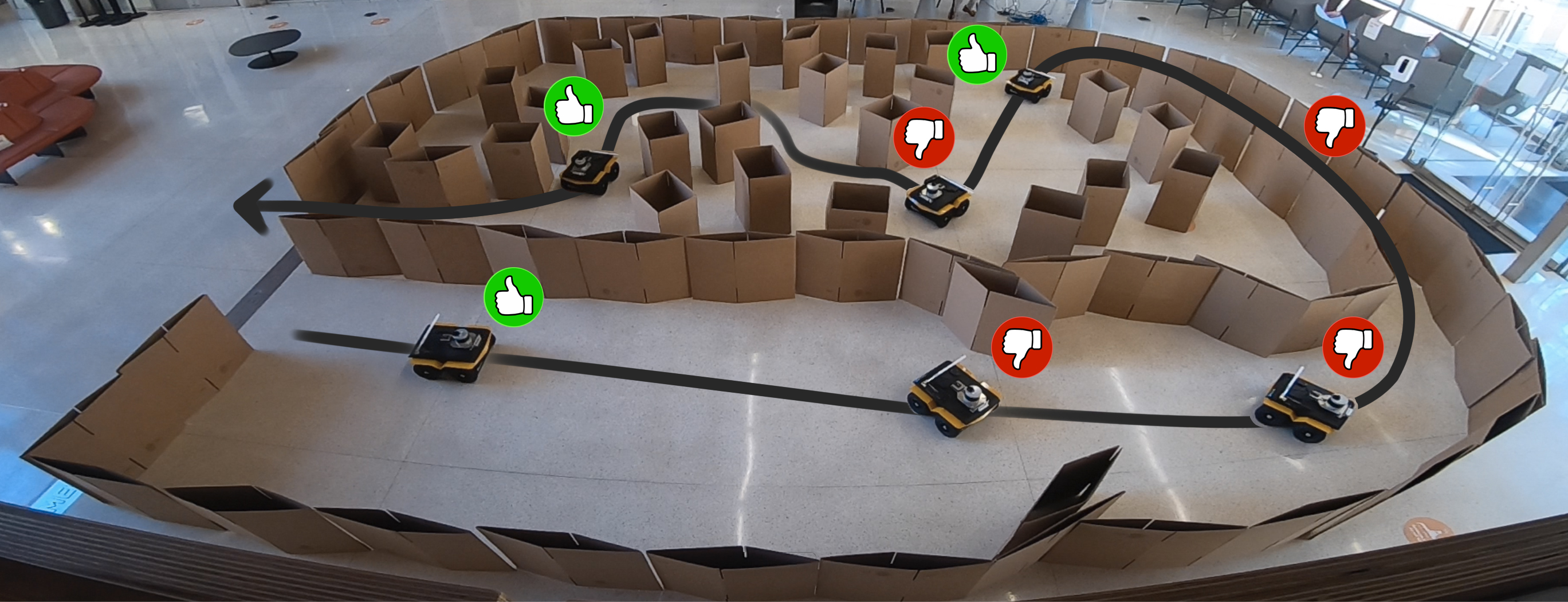}
  \caption{For non-expert users who are unable or unwilling to take control of the robot, evaluative feedback, e.g. {\em good job} (green thumbs up) or {\em bad job} (red thumbs down), is a more accessible human interaction modality, but still valuable for improving navigation systems during deployment. }
  \label{fig::apple}
\end{figure}

\textsc{appld} and \textsc{appli} require non-expert users to take full control of the moving robot with a joystick, which some non-expert users may not feel comfortable with, e.g., due to perceived risk of human error and causing collisions. For these users, we introduce \emph{Adaptive Planner Parameter Learning from Evaluative Feedback} (\textsc{apple}) \cite{wangapple}, where they only need to observe the robot navigating and provide real-time positive or negative assessments of the observed navigation behavior through {\em evaluative feedback}. This more-accessible modality provides a new interaction channel for a larger community of non-expert users with mobile robots (Fig. \ref{fig::apple}). In this section, we briefly summarize \textsc{apple} and reformalize it based on the \textsc{appl} formulation in Sec. \ref{sec::appl}. For full details and experiment results, please refer to our RAL article \cite{wangapple}. 

% \textsc{apple} has two novel features: (1) \textsc{apple} just requires evaluative feedback that can be provided even by non-expert users; (2) in contrast to previous work \cite{xiao2020appld, wang2020appli} that selects the planner parameter set based on how similar the deployment environment is to the demonstrated environment, \textsc{apple} is based on the expected evaluative feedback, i.e., the actual navigation performance.
% \textsc{apple}'s performance-based parameter policy has the potential to outperform previous approaches that are based on similarity.

\subsection{Learning \textsc{apple} Policy}
\textsc{apple} learns a parameter policy from human evaluative feedback in order to select the appropriate parameter set $\theta$ for the current deployment scenario. The parameter set can be selected from either a discrete parameter set library or from a continuous full parameter space. 
To be specific, a human observes the underlying planner taking action $a$ at state $x$, and then provides evaluative feedback $e$.
The human can provide either discrete (e.g., ``good job"/``bad job") or continuous (e.g., a score ranging in $[0, 1]$) evaluative feedback. 
With the evaluative feedback from the human, \textsc{apple} finds (1) a parameterized predictor $F_\phi: \mathcal{X} \times \Theta \rightarrow \mathcal{E}$ that predicts human evaluative feedback for each state-parameter pair $(x, \theta)$, and (2) a parameterized parameter policy $\pi_\psi: \mathcal{X} \rightarrow \Theta$ that determines the appropriate planner parameter set $\theta$ for the current state $x$.
% Based on whether \textsc{apple} chooses the parameter set from a library or the parameter space, we introduce two discrete and continuous parameter policies. 

\subsubsection{Discrete Parameter Policy}
A discrete parameter policy is designed for the situations where $K$ candidate parameter sets (e.g., the default set or sets tuned for specific environments like narrow corridors, open spaces, etc.) are already available. 
These $K$ candidate parameter sets comprise a parameter library $\mathcal{L} = \{\theta^i\}_{i=1}^K$ (superscript $i$ denotes the index in the library). In this case, a discrete \textsc{apple} policy learns to select the most appropriate of these parameters given the state $x$ using the provided evaluative feedback $e$.

In our discrete \textsc{apple} policy, the feedback predictor $F_\phi$ is parameterized in a way similar to the value network in DQN \cite{mnih2013playing}: the input is the state $x$, while the output is $K$ predicted feedback values $\{\hat{e}^i\}_{i=1}^K$. Each $\hat{e}^i$ is a prediction of the evaluative feedback a human user would give if the planner were using the respective parameter set $\theta^i \in \mathcal{L}$ at state $x$. During training, a feedback dataset for supervised learning, $\mathcal{F} = \{x_j, \theta_j, e_j\}_{j=1}^N$ ($\theta_j \in \mathcal{L}$, subscript $j$ denotes the time step), is built using the evaluative feedback collected so far. 
We use supervised learning which minimizes the difference between predicted feedback and the label to learn $F_\phi$: 
\begin{equation}
    \phi^* = \argmin_\phi \mathop{\mathbb{E}}_{(x_j, \theta_j, e_j) \sim \mathcal{F}} \ell(F_\phi(x_j, \theta_j), e_j)
    \label{eqn::critic_loss}
\end{equation}
where $\ell(\cdot, \cdot)$ is the binary cross entropy loss if the feedback $e_j$ is discrete (e.g., $e_j = 1$ for ``good job'', and $e_j = 0$ for ``bad job''),  or mean squared error given continuous feedback.

With $F_{\phi^*}$, the discrete parameter policy $\pi(\cdot|x)$ simply chooses the parameter set that maximizes the expected human feedback:  
\begin{equation}
     \pi(\cdot|x) = \argmax_{\theta \in \mathcal{L}} F_{\phi^*}(x, \theta).
     \label{eqn::discrete}
\end{equation}
Note $\psi$ is omitted since only $\phi^*$ is needed for $\pi_\psi$ to select the appropriate $\theta$. 

% When comparing with RL, especially DQN, discrete \textsc{apple} has a similar architecture and training objective. However, an important difference is that while RL optimizes future (discounted) cumulative reward, \textsc{apple} greedily maximizes the current feedback. The reason is that we assume, while supervising the robot's actions, the human will not only consider the current results but also future consequences and give the feedback accordingly. This assumption is consistent with past systems such as \textsc{tamer}~\cite{knox2009interactively}. Hence, \textsc{apple}'s greedy objective for human feedback still considers the goal of maximizing current and future performance.

\subsubsection{Continuous Parameter Policy}
\textsc{apple} can also learn to select appropriate parameters from continuous parameter spaces (e.g., deciding the max speed from $[0.1, 2]\ \mathrm{m/s}$). 
For continuous \textsc{apple} policy, the parameter policy $\pi_\psi$ and the feedback predictor $F_\phi$ are parameterized in the actor-critic style~\cite{haarnoja2018soft_application}.
Similar to the discrete \textsc{apple} policy, we also train $F_\phi$ by minimizing the difference between predicted and collected feedback using the collected dataset $ \mathcal{F} = \{x_j, \theta_j, e_j\}_{j=1}^N$ (Eqn. \ref{eqn::critic_loss}). The parameter policy $\pi_\psi$ is trained to not only choose the action that maximizes expected feedback, but also to maximize the entropy of policy $\mathcal{H}(\pi_\psi(\cdot|x))$ at state $x$. We use the same entropy regularization as Soft Actor Critic (SAC)~\cite{haarnoja2018soft_application}, so that $\pi_\psi$ favors more stochastic policies for better exploration during training:
\begin{equation}
     \psi^* = \argmin_\psi \mathop{\mathbb{E}}_{\substack{x_j \in \mathcal{F} \\ \tilde{\theta}_j \sim \pi_\psi(\cdot | x_j)}} \left[-F_\phi(x_j, \tilde{\theta}_j) + \alpha \log \pi_\psi(\tilde{\theta}_j | x_j)  \right],
     \label{eqn::continuous}
\end{equation}
where $\alpha$ is the temperature controlling the importance of the entropy bonus and is automatically tuned as in SAC~\cite{haarnoja2018soft_application}.

% \subsubsection{Deployment}
% During deployment, we measure the state $x_t$, use the parameter policy to obtain a set of parameters $\theta_t \sim \pi_\psi^*(\cdot | x_t)$ at each time step, and apply that parameter set to the navigation planner $G$.

The $LearningParameterPolicy$ subroutine in Alg. \ref{alg::appl} for \textsc{apple} to learn $\pi_E$ is simply learning $F_{\phi^*}$ with Eqn. \ref{eqn::critic_loss} and learning $\pi_{\psi^*}$ with SAC \cite{haarnoja2018soft_application} for the discrete and continuous parameter policy, respectively. 

\subsection{EXPERIMENTS}
\label{sec::experiments}
In our experiments, we find that \textsc{apple} can improve navigation performance by learning from evaluative feedback, in contrast to a teleoperated demonstration or a few corrective interventions, both of which require the non-expert user to take control of the moving robot. We also show \textsc{apple}'s generalizability to unseen environments. To be specific, we implement \textsc{apple} on the same ClearPath Jackal ground robot in \textsc{barn}~\cite{perille2020benchmarking}, and in two physical obstacle courses. We show that \textsc{apple} can achieve significant improvement compared to the default parameters, and even slightly better performance (not statistically significant) than \textsc{appli} learned from a similar obstacle course (Tab. \ref{tab::apple}).
For full details about the experimental setup and results, please refer to our RAL article \cite{wangapple}.

\begin{table}
\centering
\caption{Traversal Time in Training and Unseen Environment}
\begin{tabular}{cccc}
\toprule
 & Default & \textsc{appli} & \textsc{apple} (disc.) \\ 
\midrule
\textbf{Training} & 143.1$\pm$20.0s & 79.8$\pm$8.1s & 75.2$\pm$4.1s \\
\textbf{Unseen} & 150.5$\pm$24.0s & 86.4$\pm$1.1s & 83.9$\pm$4.6s \\
\bottomrule
\end{tabular}
\label{tab::apple}
\end{table}

\section{APPL FROM REINFORCEMENT (\textsc{APPLR})}
\label{sec::applr}
\emph{Adaptive Planner Parameter Learning from Reinforcement} (\textsc{applr}) \cite{xu2021applr} adopts the general notion of \emph{parameter policy} (Sec. \ref{sec::appl} Fig. \ref{fig::appl}).  
One disadvantage of learning planner parameters from different human interaction modalities is that the learner's performance is limited by the human's (most likely suboptimal) teleoperated demonstration, corrective interventions, and evaluative feedback. 
With reinforcement learning in a wide variety of simulation environments, \textsc{applr} does \emph{not} need access to any human interaction, and learns to make planner parameter decisions in such a way that allows the system to take suboptimal actions at one state in order to perform even better in the future. 
% and is generalizable to many deployment environments. 
% For example, while it may be suboptimal in the moment to slow down or alter the platform's trajectory before a turn, doing so may allow the system to carefully position itself so that it can go much faster in the future than if it had not. 

% Additionally, we assume going forward that a local goal $g$ is always available (as a waypoint along a coarse global path in most classical navigation systems), and we use its angle relative to the orientation of the agent, i.e., $\phi = \arctan2{(g_y, g_x)} \in [-\pi, \pi]$, to inject the local goal information to the agent.

\subsection{Learning \textsc{applr} Policy}
\textsc{applr} uses an existing RL algorithm and a reward function to learn a parameter policy $\pi_R$. 
\subsubsection{Reward Function} 
\label{sec::applr_reward}
In general, we encourage three types of behaviors: (1) behaviors that lead to the global goal faster; (2) behaviors that make more local progress; and (3) behaviors that avoid collisions and danger. Correspondingly, the designed reward function can be summarized as
\begin{equation}
    R_t(s_t, a_t, s_{t+1}) = c_fR_f + c_p R_p + c_c R_c.
\end{equation}
Here, $c_f, c_p, c_c$ are coefficients for the three types of reward functions $R_f, R_p, R_c$. 
Specifically, $R_f(s_t, a_t) = \mathbbm{1}(s_t~\text{is terminal}) - 1$ applies a $-1$ penalty to every step before reaching the global goal.
To encourage the local progress of the robot, we add a dense shaping reward $R_p$. Assume $\beta = (\beta_x, \beta_y) \in \mathbb{R}^2$ is the global goal and at time $t$, the absolute coordinates of the robot are $p_t= (p^x_t, p^y_t)$, we define
\begin{equation}
    R_p = \frac{(p_{t+1}-p_t)\cdot(\beta-p_t)}{|\beta-p_t|}, 
\end{equation}
In other words, $R_p$ denotes the robot's local progress $(p_{t+1}-p_t)$ projected on the direction toward the global goal ($\beta - p_t$). 
Finally, a penalty for the robot colliding with or coming too close to obstacles is defined as $R_c = -1/d(p_{t+1})$, where $d(p_{t+1})$ is a distance function measuring how close the robot is to obstacles based on sensor observations (e.g., using LiDAR). 

\subsubsection{Reinforcement Learning Algorithm} 
\label{sec::applr_rl}
For continuous actions and high sample efficiency, we use the Twin Delayed Deep Deterministic policy gradient algorithm (TD3) \cite{fujimoto2018addressing}, an actor-critic algorithm that keeps an estimate for both the policy and two state-action value functions.
% For better sample efficiency, we further employ a distributed general reinforcement learning architecture (Gorila) \cite{nair2015massively}, which enables parallelized acting processes on a computing cluster. 
% Our implementation of Gorila is a simpler version with only one serial learner and a large number of actors running individually in simulation environments to generate large quantities of data for a global replay buffer.

\subsection{Experiments}
\label{sec::applr_experiments}
Again, we use the same ClearPath Jackal and the same setup to validate that \textsc{applr} can enable adaptive autonomous navigation \emph{without} access to expert tuning or human demonstrations and is generalizable to many deployment environments, both in simulation and in the real-world. 
The results of \textsc{applr} are compared with those obtained by the underlying navigation system using its default parameters from the robot platform manufacturer. For the physical experiments, we also compare to parameters learned from \textsc{appld}. 

\subsubsection{Training}
\label{sec::applr_training}
$x_t$ is composed of 720-dimensional laser scan (capped to 2m) and a relative goal angle, computed from a local goal taken from the global path 1m  away from the robot. The meta-state $s_t$ is composed of $x_t$ and $\theta_{t-1}$. \textsc{applr} learns a parameter policy $\pi_R$ to select the same \textsc{dwa} parameters $\theta_t$ as other \textsc{appl} methods. 
$\pi_R$ is trained in simulation using 250 randomly selected training environments from the \textsc{barn} dataset \cite{perille2020benchmarking}. In each of the environments, the robot aims to navigate from a fixed start to a fixed goal location in a safe and fast manner. 
For the reward function, while $R_f$ penalizes each time step before reaching the global goal, we simplified $R_p$ by replacing it with its projection along the $y$-axis (longitudinal direction) of all \textsc{barn} environments, because the traversal paths from start to goal in all \textsc{barn} environments are along the positive direction of the y-axis. The distance function in $d(p_{t+1})$ in $R_c$ is the minimal value among the 720 laser beams. $\pi_R$ produces a new set of planner parameters every two seconds. 

This simulated navigation task is implemented in a Singularity container, which enables easy parallelization on a computer cluster. 
TD3 \cite{fujimoto2018addressing} is implemented to learn the parameter policy $\pi_R$ in simulation. The policy network and the two target Q-networks are represented as multilayer perceptrons with three 512-unit hidden layers. The policy is learned under the distributed architecture Gorila \cite{nair2015massively}. The acting processes are distributed over 500 CPUs with each CPU running one individual navigation task. On average, two actors work on a given navigation task. A single central learner periodically collects samples from the actors' local buffers and supplies the updated policy to each actor. Gaussian exploration noise with 0.5 standard deviation is applied to the actions at the beginning of the training. Afterward the standard deviation linearly decays at a rate of 0.125 per million steps and stays at 0.02 after 4 million steps. The entire training process takes about 6 hours and requires 5 million transition samples in total. 
% Figure \ref{fig::learning_curve} shows episode length and return averaged over 100 episodes and compares with \textsc{dwa} motion planner with default parameters. The episode return continually increases and episode length drops by around 40\% by the end of the training. As shown in Figure \ref{fig::learning_curve}, \textsc{applr} surpasses \textsc{dwa} at an early stage of the training process in terms of both episode length and return. 
% After training, we deploy the learned parameter policy $\pi_R$ in an example environment and plot four example parameter profiles produced by $\pi_R$ in Figure \ref{fig::gazebo_profile}. Dashed lines separate different parts of the profile, which correspond to different regions in the example environment. 

% \begin{figure}
%   \centering
%   \includegraphics[width=\columnwidth]{contents/figures/learning_curve.png}
%   \caption{Learning Curves of Episode Length and Episode Return Averaged over 100 Episodes: The values are moving averaged over 40k steps. The red dashed lines mark the average performance of \textsc{dwa} planner with a static set of default parameters. The curves show significant improvement in the episode return and the time steps required to finish a navigation task.}
%   \label{fig::learning_curve}
% \end{figure}

% \begin{figure*}
%   \centering
%   \includegraphics[width=2\columnwidth]{contents/figures/gazebo_profile.png}
%   \caption{Example Parameter Profiles Selected by \textsc{applr}: Labels of qualitatively similar regions are in the same color. }
%   \label{fig::gazebo_profile}
% \end{figure*}

\subsubsection{Simulated Experiments}
\begin{figure}
  \centering
  \includegraphics[width=\columnwidth]{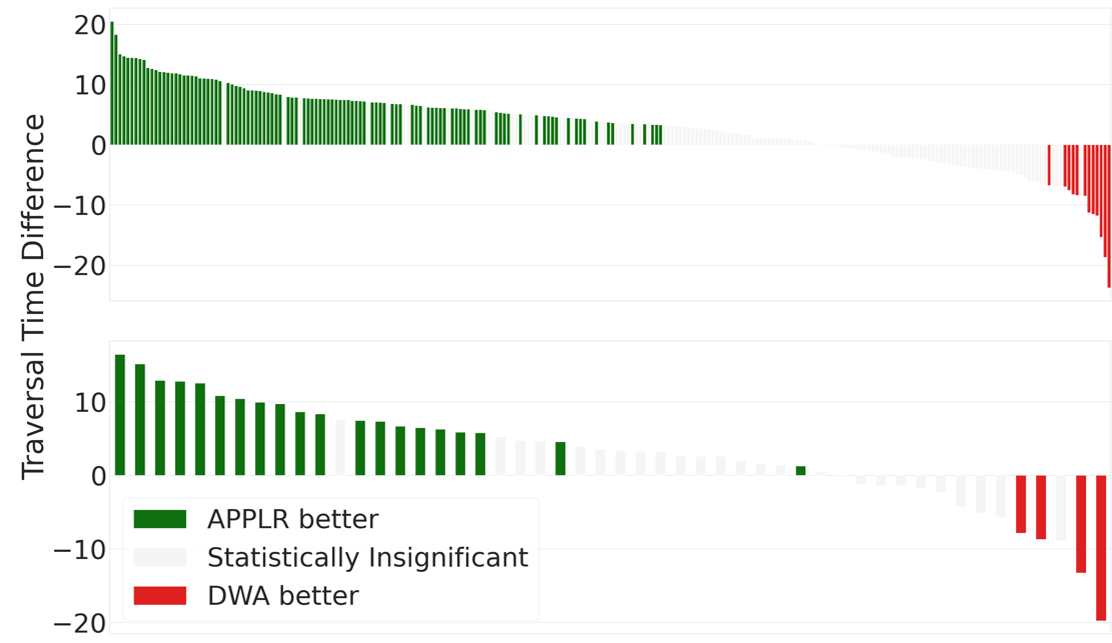}
  \caption{Traversal time difference between \textsc{applr} and \textsc{dwa} for the training environments (top) and the test environments (bottom). The bars represent the environments ordered by traversal time difference. The colored bars indicate the environments that show statistically significant difference. } 
  \label{fig::barplot_test}
\end{figure}

After training, we deploy the learned parameter policy $\pi_R$ on both the 250 training and 50 test environments. We also use traversal time to evaluate the performance of the policy. A maximum traversal time of 50s is used, and the failing trials are set to be 50s plus a 20s penalty. We average over the traversal time of 40 trials for \textsc{dwa} and \textsc{applr} in each environment. We also perform t-tests for  each  pair  of \textsc{applr} and \textsc{dwa} performance to check statistical significance. The results over the 250 training environments and 50 test environments are shown in Fig. \ref{fig::barplot_test}. For the majority of the environments (green bars), \textsc{applr} shows statistically significantly better navigation  performance. 
Tab. \ref{tab::average_epl} shows the average traversal time of \textsc{applr} and \textsc{dwa}, and relative improvement (averaged over three training runs). \textsc{applr} yields an improvement of 23.2\% in the training environments, and 14.8\% in the test environments. 
% , which corresponding to 2.56s of the real navigation time. In the test set, \textsc{applr} also decreases the episode length by 0.78.  
In addition, Tab. \ref{tab::t_test} compares the number of environments that show significant improvement and deterioration. In both training and test set, \textsc{applr} achieves statistically significantly better navigation performance in over 40\% of environments than \textsc{dwa} does, while \textsc{dwa} is only better in 4.8\% and 8\% of environments in the training and test set, respectively. 

\begin{table}
\centering
\caption{Average Traversal Time of \textsc{applr} and \textsc{dwa}}
\begin{tabular}{ccccc}
\toprule
  & \textsc{applr} & \textsc{dwa} & Improvement & P-Value \\ \midrule
Training & $20.8 \pm 1.8$ & $27.1 \pm 1.3$ & 6.3 (23.2\%)& $8.2 \times 10^{-3}$  \\ 
Test & $22.4 \pm 1.0$ & $26.3 \pm 2.1$ & 3.9 (14.8\%) & $5.3 \times 10^{-2}$ \\
\bottomrule
\end{tabular}
\label{tab::average_epl}
\end{table}

\begin{table}
\centering
\caption{Number and Percentage of All Environments 
in which One Method is Better Compared to the Other
} 
\begin{tabular}{ccccc}
\toprule
  & \textsc{applr} better & \textsc{dwa} better &   \\ \midrule
Training & 106 ($42.4\%$) & 12 ($4.8\%$) &   \\ 
Test & 20 (40\%) & 4 (8\%) &   \\
\bottomrule
\end{tabular}
\label{tab::t_test}
\end{table}

\subsubsection{Physical Experiments}
\begin{figure}
  \centering
  \includegraphics[width=\columnwidth]{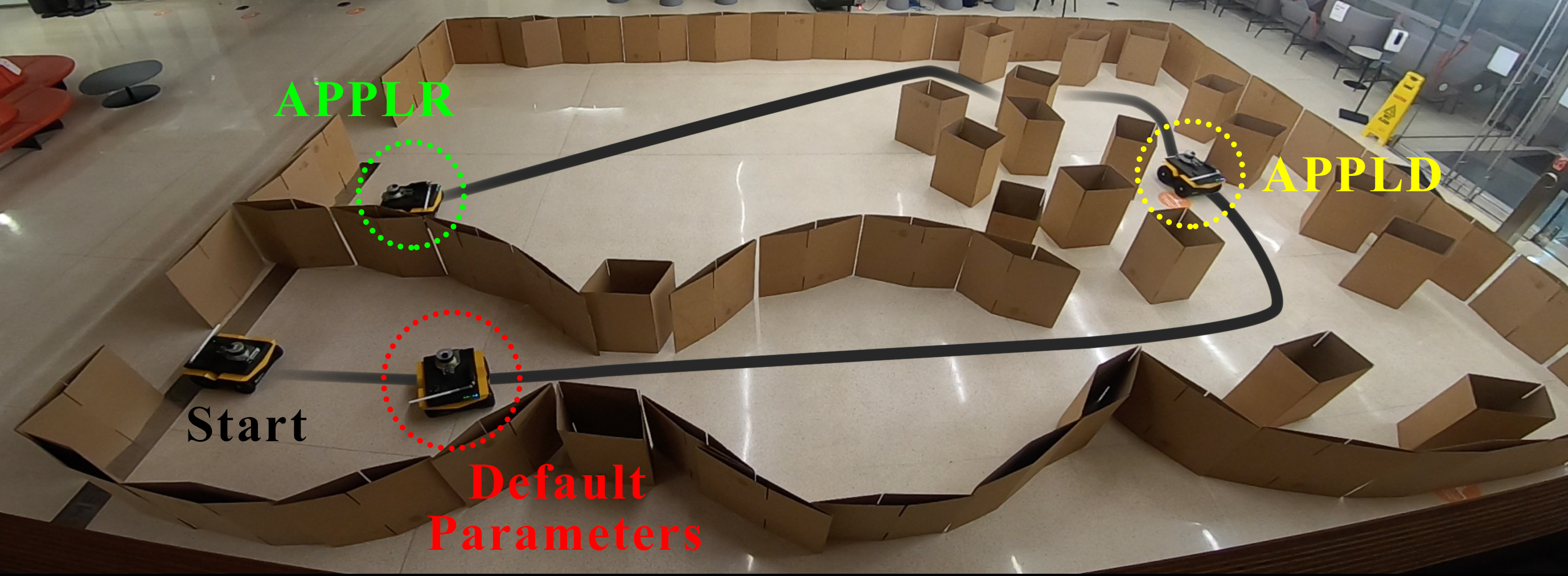}
  \caption{\textsc{applr} Physical Experiments} 
%   While the \textsc{dwa} planner with a static set of default parameters (red) fails to find feasible motions and executes recovery behaviors in many places, \textsc{appld} (yellow) and \textsc{applr} (red) can both successfully and smoothly navigate through the entire obstacle course. Using RL, \textsc{applr} can achieve faster traversal than \textsc{appld} learned from (most likely suboptimal) human demonstration. }
  \label{fig::physical}
\end{figure}

To validate the sim-to-real transfer of \textsc{applr}, we also test the learned parameter policy $\pi_R$ on a physical Jackal robot. 
% Again, we transform the 3D point cloud to the same 2D laser scan as in the simulation (720-dimensional and 270$^\circ$ field of view). 
The learned policy is deployed in a real-world obstacle course (Fig. \ref{fig::physical}). This physical environment is different from any of the navigation environments in \textsc{barn}. Therefore, both generalizability and sim-to-real transfer of \textsc{applr} can be tested with this unseen real-world environment. Note that the use of LiDAR input also reduces the difference between simulation and the real-world. 

Given this target environment, we further collect a teleoperated demonstration provided by one of the authors and learn a parameter tuning policy based on the notion of navigational context (\textsc{appld}). The author aims at driving the robot to traverse the entire obstacle course in a safe and fast manner. \textsc{appld} identifies three contexts using the human demonstration and learns three sets of navigation parameters. 

We compare the performance of \textsc{applr} with that of \textsc{appld} and the \textsc{dwa} planner using a set of hand-tuned default parameters. For each trial, the robot navigates from the fixed start point to a fixed goal point. Each trial is repeated five times and we report the mean and standard deviation in Table \ref{tab::physical_results}. 
We also conduct t-tests for \textsc{applr} vs. \textsc{dwa} and \textsc{applr} vs. \textsc{appld} and find the p-values to be $2.5 \times 10^{-4}$ and $1.4 \times 10^{-2}$ respectively, showing \textsc{applr}'s statistically significant improvement.
% We observe one failure trial (the robot fails to find feasible motions and keeps rotating in place at the beginning of the narrow part) with \textsc{dwa}. Therefore, the \textsc{dwa} results only contain the four successful trials.
In all \textsc{dwa} trials, the robot gets stuck in many places, especially where the surrounding obstacles are very tight. It has to engage in many recovery behaviors, i.e. rotating in place or driving backwards, to ``get unstuck''. Furthermore, in relatively open space, the robot drives unnecessarily slowly. All these behaviors contribute to the large traversal time and high variance (plus an additional failure trial). 
Unlike many simulation environments in \textsc{barn} \cite{perille2020benchmarking}, where obstacles are generated by cellular automata and therefore very cluttered, the relatively open space in the physical environment (Fig. \ref{fig::physical}) allows faster speed and gives \textsc{applr} a greater advantage.  
Surprisingly, \textsc{applr} even achieves better navigation performance than \textsc{appld}, which has access to a human demonstration in the same environment. One of the reasons we observe this result in the physical experiments is that the human demonstrator is relatively conservative in some places; the parameters learned by \textsc{appld} are upper-bounded by this suboptimal human performance. 
% On the other hand, the RL parameter policy aims at reducing the traversal time and finds better parameter sets to achieve that goal. 
Another reason is that \textsc{applr} is given the flexibility to continually change parameters, and RL is able to utilize the sequential aspect of the parameter selection problem, in contrast to \textsc{appld}'s three sets of static parameters. 

\begin{table}
\centering
\caption{Traversal Time in Physical Experiments \\ \protect\centering \small{(* denotes one additional failure trial)} }
\begin{tabular}{ccc}
\toprule
\textsc{dwa} & \textsc{appld} & \textsc{applr} \\ 
\midrule
% 78.2$\pm$15.0s 
72.8$\pm$10.1s* & 43.2$\pm$4.1s  & \textbf{34.4$\pm$4.8s}\\
\bottomrule
\end{tabular}
\label{tab::physical_results}
\end{table}

%%%%%%%%%%%%%%%%%%%%%%%%%%%%%%%%%%%%%%%%%%%%%%%%%%%%%%%%%%%%%%%%%%%%%%%%%%%%%%%%
\section{CYCLE-OF-LEARNING}
\label{sec::cycle}
\begin{figure*}
  \centering
  \includegraphics[width=1\columnwidth]{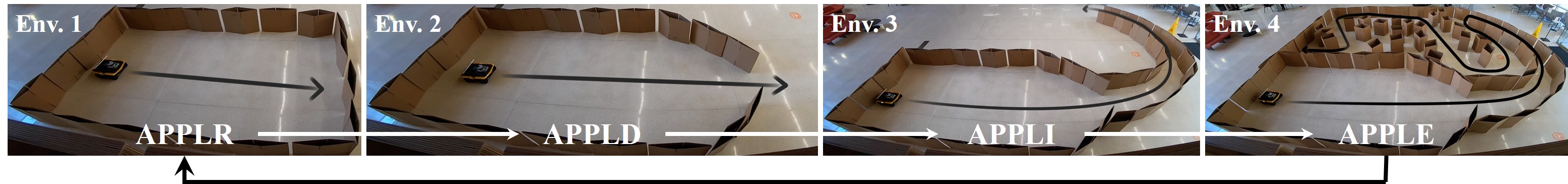}
  \caption{One Learning Cycle. White Arrows: Learning in the Real-World. Black Arrow: Learning in Simulation. }
  \label{fig::cycle}
\end{figure*}

The four individual \textsc{appl} methods are combined to form a cycle-of-learning scheme \cite{waytowich2018}, in which a mobile robot can interact with different users in different deployment environments and continually improve its navigation in a cyclic fashion. We present our cycle-of-learning algorithm in Alg. \ref{alg::cycle}. We provide detailed explanation of Alg. \ref{alg::cycle} using one full cycle (Fig. \ref{fig::cycle}) as an example, which starts from \textsc{applr}, goes through \textsc{appld}, \textsc{appli}, and \textsc{apple}, and comes back to \textsc{applr}. To distinguish the start and end of the cycle, we name them \textsc{applr1} and \textsc{applr2}. 

We assume a mobile robot is given an underlying navigation system $G$ (e.g., \textsc{dwa} planner \cite{fox1997dynamic}) and a space of possible parameters $\Theta$ (e.g., \textsc{dwa} parameters). We also assume there is a set of simulation environments, where the robot can learn through trial and error, e.g., the \textsc{barn} dataset \cite{perille2020benchmarking} (line 1 Alg. \ref{alg::cycle}). Before any real-world deployment, we first train the \textsc{applr1} policy $\pi_R$ in the \textsc{barn} dataset $\mathcal{E}$ (line 3 Alg. \ref{alg::cycle}). Then the cycle-of-learning starts (lines 5-28 Alg. \ref{alg::cycle}). With only one parameter policy $\pi_R$ in the policy set $\Pi$ (line 7 Alg. \ref{alg::cycle}), the first deployment environment in our example cycle is a relatively open space, where default \textsc{dwa} slowly drives at the default 0.5m/s max velocity and \textsc{applr1} is able to accelerate to 2.0m/s (line 9 Alg. \ref{alg::cycle}). Therefore user 0 is satisfied with the navigation performance and does not choose to interact with the robot (skipping lines 10-15 Alg. \ref{alg::cycle}). Moving on to the second environment (line 9 Alg. \ref{alg::cycle}), \textsc{applr1} produces unsmooth motion at the narrow gap (line 10 Alg. \ref{alg::cycle}). User 1 provides a full demonstration for Environment 2 (line 11 Alg. \ref{alg::cycle}), from which \textsc{appld} learns two contexts and corresponding parameters (line 12 Alg. \ref{alg::cycle}). The navigation therefore improves (line 13 Alg. \ref{alg::cycle}). When deployed in the third environment (line 9 Alg. \ref{alg::cycle}), \textsc{appld} gets stuck multiple times in the narrow corridor (line 10 Alg. \ref{alg::cycle}), where User 2 intervenes (line 11 Alg. \ref{alg::cycle}) and uses \textsc{appli} to learn an extra context (line 12 Alg. \ref{alg::cycle}). In Environment 4 (line 9 Alg. \ref{alg::cycle}), \textsc{appli} suffers from suboptimal behaviors in the obstacle field (line 10 Alg. \ref{alg::cycle}). User 3 provides evaluative feedback (line 11 Alg. \ref{alg::cycle}) and uses \textsc{apple} to improve the context predictor of \textsc{appli} (line 12 Alg. \ref{alg::cycle}). The performance of different \textsc{appl} variants in the four environments is shown in Tab. \ref{tab::cycle} (averaged over 5 trials). The method which directly utilizes human interaction in the particular environment achieves the best results in the corresponding environment (bold). 

\begin{table*}
\centering
\caption{Cycle-of-Learning Performance in Env. 1-4}
\begin{tabular}{ccccccc}
\toprule
\textbf{Env.} & \textbf{DWA}                                                  & \textbf{APPLR1}                                                & \textbf{APPLD}                                               & \textbf{APPLI}                                               & \textbf{APPLE}                                               & \textbf{APPLR2}                                              \\
\midrule
\textbf{1}    & \begin{tabular}[c]{@{}c@{}}9.8\\ $\pm$0.5s\end{tabular}   & \begin{tabular}[c]{@{}c@{}}\textbf{4.4}\\ \textbf{$\pm$0.3s}\end{tabular}    &                                                     &                                                     &                                                     &                                                     \\
\textbf{2}    & \begin{tabular}[c]{@{}c@{}}30.2\\ $\pm$1.4s\end{tabular}  & \begin{tabular}[c]{@{}c@{}}18.6\\ $\pm$3.6s\end{tabular}   & \begin{tabular}[c]{@{}c@{}}\textbf{12.5}\\ \textbf{$\pm$0.4s}\end{tabular} &                                                     &                                                     &                                                     \\
\textbf{3}    & \begin{tabular}[c]{@{}c@{}}56.9\\ $\pm$2.5s\end{tabular}  & \begin{tabular}[c]{@{}c@{}}57.5\\ $\pm$5.8s\end{tabular}   & \begin{tabular}[c]{@{}c@{}}38.1\\ $\pm$2.3s\end{tabular} & \begin{tabular}[c]{@{}c@{}}\textbf{37.0}\\ \textbf{$\pm$2.0s}\end{tabular} &                                                     &                                                     \\
\textbf{4}    & \begin{tabular}[c]{@{}c@{}}109.2\\ $\pm$6.5s\end{tabular} & \begin{tabular}[c]{@{}c@{}}103.4\\ $\pm$12.8s\end{tabular} & \begin{tabular}[c]{@{}c@{}}90.6\\ $\pm$6.0s\end{tabular} & \begin{tabular}[c]{@{}c@{}}88.8\\ $\pm$2.9s\end{tabular} & \begin{tabular}[c]{@{}c@{}}\textbf{76.4}\\ \textbf{$\pm$3.5s}\end{tabular} & \begin{tabular}[c]{@{}c@{}}81.6\\ $\pm$5.6s\end{tabular}\\
\bottomrule
\end{tabular}
\label{tab::cycle}
\end{table*}

We further perform t-tests for the best two methods in each of the four environments and report the corresponding p-values in Table \ref{tab::cycle-t-test}. In Environments 1, 2, and 4, \textsc{applr1}, \textsc{appld}, and \textsc{apple} are statistically significantly better than the default \textsc{dwa}, \textsc{applr1}, and \textsc{appli}, respectively. In Environment 3, \textsc{appli} does not achieve statistically significant improvement over \textsc{appld}, possibly because the extremely difficult extra narrow corridor in Environment 3 also causes trouble for the parameters learned by \textsc{appli}. 

\begin{table}
\centering
\caption{P-Values of the Best Two Methods for Env. 1-4}
\begin{tabular}{ccccc}
\toprule
& \textbf{\begin{tabular}[c]{@{}c@{}}Env. 1\\ R1 vs. DWA\end{tabular}} & \textbf{\begin{tabular}[c]{@{}c@{}}Env. 2\\ D vs. R1\end{tabular}} & \textbf{\begin{tabular}[c]{@{}c@{}}Env. 3\\ I vs. D\end{tabular}} & \textbf{\begin{tabular}[c]{@{}c@{}}Env. 4\\ E vs. I\end{tabular}} \\
\midrule
p-value & $4.5 \times 10^{-8}$ & $5.5 \times 10^{-3}$ & $0.45$   & $2.9 \times 10^{-4}$\\
\bottomrule
\end{tabular}
\label{tab::cycle-t-test}
\end{table}

After deploying in Environment 4 with \textsc{apple}, we finish all physical deployment in this cycle (line 16 Alg. \ref{alg::cycle}) and then train \textsc{applr2} on the \textsc{barn} dataset (lines 18-27 Alg. \ref{alg::cycle}). For each policy learned from different human interactions in the current policy set $\Pi$ (line 19 Alg. \ref{alg::cycle}), we identify the \textsc{barn} environments where the \textsc{appl} variants perform better than \textsc{applr1} (line 21 Alg. \ref{alg::cycle}), and use the learned \textsc{appld}, \textsc{appli}, or \textsc{apple} policies for exploration (line 22 Alg. \ref{alg::cycle})\footnote{Other techniques for combining policies may be possible (e.g., \cite{goecks2020integrating}).} with a probability which decreases from 1 to 0 in those environments. \textsc{applr2} achieves significantly better performance in these environments due to the better exploration policy learned within the cycle from different human interaction modalities. We further test \textsc{applr2}'s performance against \textsc{applr1}'s in all \textsc{barn} environments and do not observe significant deterioration in other environments. In fact, \textsc{applr2}'s average traversal time in \textsc{barn} improves from \textsc{applr1}'s 24.7s to 24.0s. We also deploy \textsc{applr2} in Environment 4 in Fig. \ref{fig::cycle}. Although the training experience in \textsc{barn} slightly increases traversal time compared to \textsc{apple}, \textsc{applr2} significantly outperforms \textsc{applr1} by incorporating all human interactions, including the learned context predictor and parameters, into a stand-alone parameter policy. We hypothesize that \textsc{applr2}'s performance degradation compared to \textsc{apple} in Environment 4 arises because while \textsc{apple}'s feedback specifically targets improving performance in Environment 4, \textsc{applr2} has to consider performance in the simulated \textsc{barn} environments as well, which may compromise the performance in Environment 4.

% \begin{algorithm}
% \caption{\textsc{Cycle-of-Learning}} \label{alg::cycle}
% \begin{algorithmic}[1]
%     \STATE{\textbf{Input:} \textsc{applr} policy $\pi_R$, \textsc{appld} policy $\pi_D$, \textsc{appli} policy $\pi_I$, \textsc{apple} policy $\pi_E$, and simulation environments $\mathcal{E}$}.
%     \STATE Initialize training dataset $\mathcal{D}=\emptyset$.
%     \STATE Initialize parameter policy with $\pi_R$.
%     \FOR{$\pi \in \{\pi_D, \pi_I, \pi_E\}$}
%         \FOR{$e_i \in \mathcal{E}$}
%             \IF{$\pi$ performs better than $\pi_r$ in $e_i$}
%                 \STATE{Explore with $\pi$ in $e_i$ to generate training data $d$.}
%                 \STATE $\mathcal{D} = \mathcal{D} \cup d$.
%             \ENDIF
%         \ENDFOR
%     \ENDFOR
%     \STATE Update $\pi_R$ with $\mathcal{D}$ for next cycle of physical deployments. 
% \end{algorithmic}
% \end{algorithm}

\begin{algorithm}
\caption{\textsc{Cycle-of-Learning}} \label{alg::cycle}
\begin{algorithmic}[1]
    \STATE{\textbf{Input:} navigation stack $G$, space of possible parameters $\Theta$, and simulation environments $\mathcal{E}$}.
    \STATE{// Initialization}
    \STATE Train \textsc{applr} policy $\pi_R$ in $\mathcal{E}$ to select $\theta \in \Theta$ for $G$.
    \STATE{// Cycle-of-Learning}
    \WHILE{True}
        \STATE{// Physical Deployment}
        \STATE Initialize policy set $\Pi=\{\pi_R\}$.
        \FOR{each deployment}
            \STATE Deploy a policy $\pi \in \Pi$ (e.g., the latest one).
            \IF{unsatisfactory navigation performance}
                \STATE User provides interaction $\mathcal{I}$ (Demonstration, Interventions, or Evaluative Feedback).  
                \STATE $\pi = LearnParameterPolicy(\mathcal{I}, \Theta, G)$ (\textsc{appld}, \textsc{appli}, or \textsc{apple}, based on the chosen modality).
                \STATE Deploy with $\pi$.
                \STATE $\Pi = \Pi \cup \pi$. 
            \ENDIF
        \ENDFOR
        \STATE{// Simulated Training}
        \STATE Initialize training dataset $\mathcal{D}=\emptyset$.
        \FOR{each $\pi \in \Pi \setminus \pi_R$}
            \FOR{each $e_i \in \mathcal{E}$}
                \IF{$\pi$ performs better than $\pi_R$ in $e_i$}
                    \STATE{Explore with $\pi$ in $e_i$ to gather training data $d$.}
                    \STATE $\mathcal{D} = \mathcal{D} \cup d$.
                \ENDIF
            \ENDFOR
        \ENDFOR
        \STATE Update $\pi_R$ with $\mathcal{D}$ for next cycle of deployments. 
    \ENDWHILE
\end{algorithmic}
\end{algorithm}

% \begin{figure}
%   \centering
%   \includegraphics[width=\columnwidth]{contents/figures/applr2v1.png}
%   \caption{Traversal time difference between \textsc{applr2} and \textsc{applr1}. } 
%   \label{fig::applr2v1}
% \end{figure}
%%%%%%%%%%%%%%%%%%%%%%%%%%%%%%%%%%%%%%%%%%%%%%%%%%%%%%%%%%%%%%%%%%%%%%%%%%%%%%%%
\section{DISCUSSION}
\label{sec::discussions}
The \textsc{appl} paradigm integrated with the Cycle-of-Learning scheme opens up at least three dimensions for autonomous mobile robots deployed in the real-world co-existing with non-expert human users. First, building upon classical navigation planners, the \textsc{appl} agent interfaces with these systems through their hyper-parameters. Therefore, benefits of these classical systems can be inherited when being deployed in the real world, e.g., safety can still be provided by the provable guarantees of the classical systems and the behavior of the \textsc{appl} agent is still explainable through the designed role of each learned hyper-parameter. While enjoying these benefits, \textsc{appl} simultaneously empowers mobile robots with adaptivity to a variety of scenarios in the wild, which is usually an advantage of purely learning-based methods.

Second, \textsc{appl} allows mobile robots to learn from a variety of non-expert human users through multi-model interactions, ranging from teleoperated demonstration, corrective interventions, and evaluative feedback. Relaxing the assumption of access to expert roboticists when facing new deployment scenarios or suboptimal navigation behavior in existing environments, many non-expert human users, or even bystanders, can also contribute to the improvement of autonomous navigation performance in the real world.

Third, the Cycle-of-Learning scheme further removes the traditionally myopic focus on in-situ adjustment or improvement in one single deployment scenario with a particular human interaction, and aims toward a human-robot ecosystem where with non-expert humans' help robots are able to continually improve during real-world deployment throughout their lifetime. Robots are then expected to require less and less frequent interactions with human users but achieve better and better performance in their future deployments.

Such a learning paradigm that leverages the agent's own exploration experiences and interactions with other agents, human or artificial, has been considered by the learning community for many years. Lin \cite{lin1992self} proposed to combine reinforcement learning with teaching frameworks to speed up reinforcement learning in solving complicated learning tasks three decades ago. At that time, the complicated learning task only involved navigating in a discrete 2D maze environment, similar to an Atari game. A decade later, Smart and Kaelbling \cite{smart2002effective} proposed a similar paradigm and moved closer toward real-world robotics applications. Similar to \textsc{appl}'s underlying classical planner, their method utilized a supplied control policy (either actual control code, or a human directly controlling the robot with a joystick) to kick start the first learning phase, where the learning agent only passively watches the supplied policy generating states, actions, and rewards and learns a value function. In the second phase, the learning agent takes control of the robot using the learned value function from the first phase. Such a paradigm assumes that after the first phase, the learning agent is fully capable of reliably executing any task in the workspace, which, however, is not always true. During real-world robot deployment which is the focus of \textsc{appl}, encountering unseen scenarios is inevitable, and the learning agent may find it difficult to generalize well to such scenarios. On the other hand, a classical scripted system or a human may be able to address these scenarios in a more trustworthy way.

The \textsc{appl} paradigm with the Cycle-of-Learning scheme is very reminiscent of the approach proposed by Smart and Kaelbling \cite{smart2002effective}, i.e., leveraging classical non-learning-based methods and/or humans to improve learning, but with a focus on addressing the generalizability issue in real-world unseen scenarios mentioned above. We posit that a robot interacting with the real-world with an end-to-end learning system will inevitably encounter situations which it cannot handle, therefore \textsc{appl} relies on classical systems throughout the entire robot deployment period to assure safety and explainability, in contrast to only during the initial phase. The learning only happens at the hyper-parameter level to fine-tine the underlying classical planner to address different deployment scenarios.

Furthermore, while Lin's work \cite{lin1992self} proposed to use teaching to accelerate reinforcement learning of complex tasks, the learning overhead of training has been largely alleviated by the development of faster computation hardware, e.g., GPUs, and better function approximators, e.g., deep neural networks, in the last three decades. Leveraging the reduced learning cost, \textsc{appl} can train a parameter policy in a variety of simulation environments, i.e., the \textsc{barn} dataset, with a highly parallelizable, containerized, distributed training system \cite{nair2021using}, in order to cover as many real-world environments as possible in simulation. During real-world deployment, when out-of-distribution scenarios are inevitably encountered, non-expert users can help the robots to adapt to these unseen scenarios using selected interaction modalities. Although it is beyond the scope of this paper, another computational factor is onboard computation \cite{dromnelle2020reduce, dromnelle2020coping} for the multi-model learning during deployment. Due to limited onboard resources, such learning may not take place on the robot, and may need to be uploaded to offboard computation resources to assure the parameter policy can be updated quickly and transferred back to the robot during a particular deployment. How to allocate both onboard and offboard computation and how to balance the trade-off between high computation power and communication latency remains an interesting future research direction.

%%%%%%%%%%%%%%%%%%%%%%%%%%%%%%%%%%%%%%%%%%%%%%%%%%%%%%%%%%%%%%%%%%%%%%%%%%%%%%%%
\section{CONCLUSIONS}
\label{sec::conclusions}
We present \textsc{appl}, a learning paradigm that leverages different human interaction modalities, including demonstration, interventions, evaluative feedback, and unsupervised reinforcement learning, to dynamically fine-tune classical navigation planner parameters to achieve better navigation performance. In addition to the individual \textsc{appl} methods and experiments, we also introduce a cycle-of-learning scheme, in which different human interaction modalities can be utilized to continually improve future navigation performance in a cyclic fashion. 
% \xuesu{Peter: We need a paragraph with limitations and future work.}
One interesting direction for future work is to investigate other methods to incorporate human interaction to RL training in simulation, e.g., using inverse reinforcement learning to infer an underlying reward function, or creating high fidelity simulations similar to real-world deployment environments.

\section*{ACKNOWLEDGMENTS}
This work has taken place in the Learning Agents Research Group (LARG) at UT Austin. LARG research is supported in part by NSF (CPS-1739964, IIS-1724157, NRI-1925082), ONR (N00014-18-2243), FLI (RFP2-000), ARO (W911NF19-2-0333), DARPA, Lockheed Martin, GM, and Bosch. Peter Stone serves as the Executive Director of Sony AI America and receives financial compensation for this work. The terms of this arrangement have been reviewed and approved by the University of Texas at Austin in accordance with its policy on objectivity in research.
%%%%%%%%%%%%%%%%%%%%%%%%%%%%%%%%%%%%%%%%%%%%%%%%%%%%%%%%%%%%%%%%%%%%%%%%%%%%%%%%

\bibliographystyle{elsarticle-num}
\bibliography{references.bib}

\end{document}